\documentclass[10pt,journal,compsoc]{IEEEtran}
\usepackage[utf8]{inputenc}
\usepackage{bm}
\usepackage{tabularx}
\usepackage{graphicx}
\usepackage{amssymb}
\usepackage{setspace}
\usepackage{amsmath}
\usepackage{array, multirow}
\usepackage{diagbox}
\usepackage[ruled,vlined]{algorithm2e}
\usepackage{tabto}
\usepackage{hyperref}

\ifCLASSOPTIONcompsoc
  \usepackage[nocompress]{cite}
\else
  \usepackage{cite}
\fi
\ifCLASSINFOpdf
\else
\fi

\newcolumntype{P}[1]{>{\centering\arraybackslash}p{#1}}
\makeatletter
\def\thickhline{%
  \noalign{\ifnum0=`}\fi\hrule \@height \thickarrayrulewidth \futurelet
   \reserved@a\@xthickhline}
\def\@xthickhline{\ifx\reserved@a\thickhline
               \vskip\doublerulesep
               \vskip-\thickarrayrulewidth
             \fi
      \ifnum0=`{\fi}}
\makeatother

\newlength{\thickarrayrulewidth}
\makeatletter
\algocf@newcmdside@kobe{LetIn@let}{%
    \KwSty{EM Optimization:} repeat until convergence:%
    \ifArgumentEmpty{#1}\relax{ #1}%
    \algocf@group{#2}%
    \par
}
\newcommand\EMOpti[1]{%
    \LetIn@let{#1}%
}

\algocf@newcmdside@kobe{EM@EStep}{%
    \KwSty{E-Step:} Compute matrix $P$:%
    \ifArgumentEmpty{#1}\relax{ #1}%
    \algocf@group{#2}%
    \par
}
\newcommand\EStep[1]{%
    \EM@EStep{#1}%
}

\algocf@newcmdside@kobe{EM@MStep}{%
    \KwSty{M-Step:} Solve for optimal transformation $\mathcal{T}$%
    \ifArgumentEmpty{#1}\relax{ #1}%
    \algocf@group{#2}%
    \par
}
\newcommand\MStep[1]{%
    \EM@MStep{#1}%
}
\makeatother

\newcommand{\vect}[1]{\bm{#1}} 

\begin{document}
%

\title{Registering Image Volumes using 3D SIFT \\ and Discrete SP-Symmetry}
%
%

%
%
\author{Laurent~Chauvin, William~Wells~III and Matthew~Toews
\thanks{L. Chauvin and M. Toews are with the Ecole de Technologie Superieure, Montreal, QC H3C 1K3, Canada (email: laurent.chauvin.2@etsmtl.net matt.toews@gmail.com).}
\thanks{W. Wells is with the Harvard Medical School, Boston, MA 02115, USA and the Massachusetts Institute of Technology, Boston, MA 02139, USA (email: sw@bwh.harvard.edu).}
\thanks{Manuscript submitted May 15, 2022. }}
%
%

\markboth{Journal of \LaTeX\ Class Files,~Vol.~14, No.~8, May~2022}%
{Chauvin \MakeLowercase{\textit{et al.}}: Image Registration based on 3D Feature Properties}
%



\IEEEtitleabstractindextext{%
\begin{abstract}
This paper proposes to extend local image features in 3D to include invariance to discrete symmetry including inversion of spatial axes and image contrast. A binary feature sign $s \in \{-1,+1\}$ is defined as the sign of the Laplacian operator $\nabla^2$, and used to obtain a descriptor that is invariant to image sign inversion $s \rightarrow -s$ and 3D parity transforms $(x,y,z)\rightarrow(-x,-y,-z)$, i.e. SP-invariant or SP-symmetric. SP-symmetry applies to arbitrary scalar image fields $I: R^3 \rightarrow R^1$ mapping 3D coordinates $(x,y,z) \in R^3$  to scalar intensity $I(x,y,z) \in R^1$, generalizing the well-known charge conjugation and parity symmetry (CP-symmetry) applying to elementary charged particles. Feature orientation is modeled as a set of discrete states corresponding to potential axis reflections, independently of image contrast inversion. Two primary axis vectors are derived from image observations and potentially subject to reflection, and a third axis is an axial vector defined by the right-hand rule. Augmenting local feature properties with sign in addition to standard (location, scale, orientation) geometry leads to descriptors that are invariant to coordinate reflections and intensity contrast inversion. Feature properties are factored in to probabilistic point-based registration as symmetric kernels, based on a model of binary feature correspondence. Experiments using the well-known coherent point drift (CPD) algorithm demonstrate that SIFT-CPD kernels achieve the most accurate and rapid registration of the human brain and CT chest, including multiple MRI modalities of differing intensity contrast, and abnormal local variations such as tumors or occlusions. SIFT-CPD image registration is invariant to global scaling, rotation and translation and image intensity inversions of the input data.
\end{abstract}

}

\maketitle

\IEEEdisplaynontitleabstractindextext

%
\IEEEpeerreviewmaketitle

\IEEEraisesectionheading{\section{Introduction}\label{sec:introduction}}

%
%
%
%

3D structure is often described in terms of localized phenomena or features. In classical physics, fine-scale matter in the atomic model is characterized via the Schrodinger equation~\cite{schrodinger1926undulatory}, where the symmetry of the Laplacian operator ensures that atomic properties remain invariant to changes in the viewpoint of the observer. Similarly, elementary 3D image features may be identified in generic image volumes via fundamental mathematical operators, e.g. the 3D Laplacian-of-Gaussian operator~\cite{Toews2013a,rister2017volumetric}, following from scale-space theory and the 2D scale-invariant feature transform (SIFT)~\cite{Lindeberg1998,Lowe2004DistinctiveKeypoints}. These are used in highly robust and efficient memory-based applications, e.g. detection~\cite{flitton2010object}, registration~\cite{Machado2018Non-rigidMatching,luo2018feature}, segmentation~\cite{wachinger2018keypoint,gill2014robust}, indexing~\cite{chauvin2021efficient,Toews2015} in a manner invariant to transformations including translation, rotation and scaling, and are applicable in generic contexts with no training or calibration procedures.

Our work here seeks to advance the understanding of 3D keypoints observed in image volumes from the notion of symmetry, specifically discrete symmetry in feature orientation. A geometrical object may be said to exhibit a symmetry if its observed properties remain unaffected by (or are invariant to) a group of transforms, i.e., a symmetry group. In particle physics, discrete symmetries include axis reflections and also simultaneous charge conjugation (inversion), 3D parity transformation and time reversal (CPT-symmetry). The CPT transform models the transition of a charge-associated particle to its anti-particle, e.g. a proton to an anti-proton, and is currently a topic of high interest~\cite{borchert202216,Lehnert2022} as symmetry violations have been observed in special cases including charge conjugation and parity (CP) transforms~\cite{christenson1964evidence,sozzi2008discrete}. Nevertheless in image analysis, discrete shape symmetries including reflections~\cite{liu2010computational,cohen2019gauge} and image contrast inversion~\cite{chen2009real,hossain2012effective,Teng2015MultimodalDescriptors,Alexander2001,Lv2019} are typically represented as separate, unrelated phenomena, typically in 2D coordinate space where 3D parity does not apply. Indeed, image processing and graphics research focuses largely on representations and challenges stemming from 2D data, including photographs, surface meshes or projections, and achieving invariance to viewpoint changes. Despite the compelling analogy between local features and the atomic model, the notion of a local feature sign, analogous to a particle charge, has not been proposed nor linked to the 3D parity transform in the context of image processing.
\begin{figure*}[ht]
    \centering
    \includegraphics[width=0.85\textwidth]{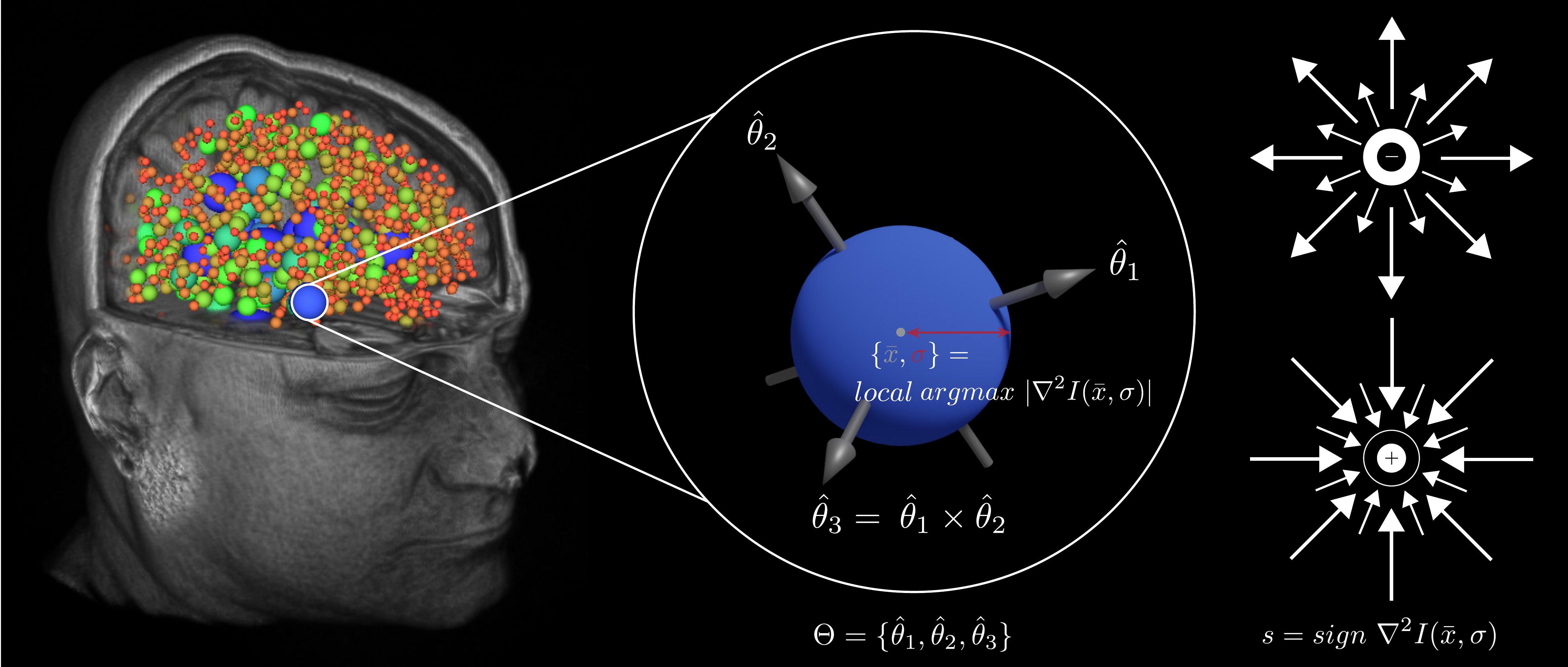}
    \caption{Illustrating the geometrical properties of 3D features (colored spheres) extracted from an MRI image of the human brain. Properties include location $\vect{x}$, orientation $\Theta$, scale $\sigma$, we introduce a binary feature sign $s = \{-,+\}$ based on the sign of the Laplacian-of-Gaussian.}
    \label{fig:properties}
\end{figure*}

Our primary contribution is the first model of discrete symmetry including intensity inversion in the context of local feature processing as shown in Figure~\ref{fig:properties}. We note that a CP-like transform occurs in the context of image processing, where a local feature observed in different imaging modalities may exhibit simultaneous contrast inversion and gradient reversal. We refer to this as a sign inversion and parity transform (SP-transform), which applies generally to arbitrary scalar image fields $I: R^3 \rightarrow R^1$ mapping 3D coordinates $(x,y,z) \in R^3$ to a scalar image intensity $I(x,y,z) \in R^1$, and we seek SP-invariance. Local features are endowed with a binary sign $s \in \{-1,+1\}$ and a set of discrete orientation states, allowing feature descriptors and image registration to be computed in a manner invariant to reflections and intensity contrast inversions, thus achieving SP-invariance. Volumetric data allows physical matter and local image features to be observed in situ within isotropic 3D space and free from projective distortion, thus facilitating the analogy between local image features and atoms in terms of charge. Local feature orientation is defined by a 3D rotation matrix or coordinate reference frame, where two primary axes are estimated from dominant image gradient directions and the third vector is defined by the cross product, ensuring feature reference frames are restricted to $SO(3)$ despite potential parity transform due to contrast inversion. Four discrete orientation states are defined, corresponding to inversions of the two primary axis vectors, due to a combination of symmetric image gradient patterns and intensity contrast inversion. Gradient-based descriptors may be inverted via sign change, independently of orientation state.

Our secondary contribution is to incorporate invariant feature properties into a probabilistic point-based registration framework. The variability of feature scale, orientation and location between images is modeled via an exponential kernel function, approximating a zero-mean noise model, invariant to feature reflection and intensity contrast inversion. Experiments adapt our model to the well-known Coherent Point Drift (CPD)~\cite{myronenko2010point} algorithm, and demonstrate that our SIFT-CPD model of enhanced feature geometry leads to faster and more accurate inter-subject registration, in a diverse variety of image data including multiple T1 and T2 weighted MRI modalities of the human brain, tumors and CT images of the human torso. We also demonstrate that our model may be used to achieve SP-invariance independently of the method used to estimate feature orientation, e.g. via principal component analysis or maximum gradient directions.

The remainder of this paper describes related work, method and experiments. Throughout the paper, we aim to provide an accessible yet complete presentation of related concepts in 3D geometry, image processing and particle physics.

\section{Related work}

Our work focuses on characterizing 3D features extracted from volumetric images, and extending invariance to account for discrete symmetry due to both axis reflection and intensity contrast inversion.

\subsection{Local Image Features}


In computer vision and image processing, informative image structures are typically identified via mathematically-defined filters or convolution operators. First order gradient $\nabla I$ operators~\cite{harris1988combined,shi1994good} may be used to identify corner-like structures, in a manner invariant to translation and rotation. Second-order derivative operators such as the Laplacian $\nabla^2 I$ may be used to identify blob-like structures, and to extend feature invariance to scale or size based on Gaussian scale-space theory\cite{Lindeberg1998}, as in the seminal scale-invariant feature transform (SIFT)~\cite{Lowe2004DistinctiveKeypoints}. The gradient and Laplacian operators may be regarded as minimal filter representations required to localize singular patterns in a scalar image $I: \mathcal{R}^D\rightarrow\mathcal{R}^1$ defined over a $D$-dimensional coordinate space. Extended representations may be used to achieve affine invariance~\cite{Mikolajczyk2005}, or make use of alternative representations, eg. non-linear scale spaces~\cite{alcantarilla2012kaze}, information theory~\cite{kadir2001saliency,toews2010mutual}, integral images for efficiency\cite{Bay2008}. Over-complete filters banks, ie. convolutional neural networks (CNNs), may be trained for the purpose of identifying point correspondences~\cite{jiang2021cotr,ono2018lf,noh2017large}, possibly from training labels generated from fundamental operators~\cite{Yi2016LIFT:Transform,Detone2018SuperPoint:Description}.


Local features incorporate descriptors of the image content in a region or receptive field. The Gaussian scale-space~\cite{Lindeberg1998} may be viewed as a CNN with a single Gaussian channel per layer~\cite{carluer2021gpu}, where the scale or receptive field size is defined explicitly by the Gaussian standard deviation $\sigma$, and is sampled in 3 logarithmic increments between $[\sigma,2\sigma]$ prior to subsampling. The receptive fields of deep network activations are a function of the filter sizes and depth~\cite{luo2016understanding}, and may be relatively large and dependent on the image contents due to non-linear operations including max pooling and ReLu~\cite{noh2017large}. Traditional descriptors have made use of gradients $\nabla I$, including gradient orientation histograms SIFT~\cite{Lowe2004DistinctiveKeypoints}, binary comparisons~\cite{calonder2011brief} and variants such as rank ordering~\cite{Toews2009}, RootSIFT~\cite{arandjelovic2012three}. Descriptors may be learned via triplet loss and gradient analysis~\cite{ono2018lf,tyszkiewicz2020disk}. While GPU-based deep learning is the defacto state-of-the-art for image classification~\cite{Krizhevsky2012}, variants of the traditional gradient representations remain competitive with learned variants~\cite{Balntas2017HPatches:Descriptors, Schonberger2017ComparativeFeatures}, specifically for matching images of non-planar objects~\cite{moreels2007evaluation} and image retrieval~\cite{bellavia2020there}.

The SIFT algorithm has been generalized to 3D volumetric images. A body of work investigates SIFT-like detectors in 3D video coordinates $(x,y,t)$ defined by 2D space $x,y$ and 1D time $t$~\cite{laptev2005space,scovanner20073}. In the context of 3D space $x,y,z$, applications include object detection~\cite{flitton2010object}, medical image analysis~\cite{cheung2009n,allaire2008full,Toews2013a,rister2017volumetric}, segmentation~\cite{gill2014robust,wachinger2018keypoint}, alignment~\cite{bersvendsen2016robust}, image stitching~\cite{ni2008volumetric}, large-scale indexing~\cite{Toews2015}, population studies~\cite{Toews2016,Kumar2018}. Most recently, the Jaccard distance between feature sets was introduced to automatically flag errors in large public training MRI datasets~\cite{Chauvin2019AnalyzingManifold,Chauvin2020NeuroimageRelatives}, and to identify family members from brain MRI~\cite{chauvin2021efficient}, capabilities facilitated by highly efficient feature indexing. We adopt the 3D SIFT-Rank method~\cite{Toews2013a} using a GPU-optimized implementation~\cite{carluer2021gpu} first used in the context of brain MRI analysis~\cite{pepin2020large}. 


\subsection{Image Registration}
Registration is a fundamental image processing task, and seeks to identify a coordinate transform $T: \Omega_1 \rightarrow \Omega_2$ between the coordinate systems $\Omega_1 \in R^3$, $\Omega_2 \in R^3$ of the same object or scene, often in three spatial dimensions, based on a pair of observed images $(I_1$,$I_2)$. Registration generally makes use of intensity and geometry information~\cite{Saiti2020AnMethods}, our work focuses on properties of local feature geometry, specifically location, orientation, scale and sign.

Point-based registration approaches are most generally applicable to local features, where the loss function minimizes the distance between pairs of points or points and a model. Examples of point cloud registration algorithms~\cite{pomerleau2015review} include iterative algorithms such as the Iterative Closest Point (ICP) algorithm~\cite{besl1992method} where points contribute uniformly to a solution, or the Coherent Point Drift (CPD) algorithm~\cite{myronenko2010point} where points contribute according to a probabilistic weighting. 3D SIFT keypoint correspondences have been used to achieve point-based registration the context of image-guided neurosurgery~\cite{luo2018feature}, including non-rigid image registration via thin plate splines~\cite{machado2018non}, finite element methods\cite{frisken2019preliminary} and the CPD algorithm~\cite{Jiao2019A3D-SIFT}, however these have made use of keypoint locations and not orientation and scale properties as we propose.

Feature-based registration methods generally consider points with properties beyond simple location. Points on a surface model may be endowed with properties such as mass~\cite{Golyanik2016GravitationalRegistration} or charge~\cite{Jauer2019EfficientClouds}, however these properties tend to be assigned algorithmically, eg. uniformly assigned across points, and not derived from the image content itself. Local properties including orientation, scale and affine deformation may be used to enhance point-based registration~\cite{riggi2006fundamental,Ma2017RemoteMatching}. Smooth deformations may be computed in a manner consistent with local rigid or affine reference frames~\cite{arsigny2009fast}. The variability of geometrical misalignment is related to feature scale $\sigma$~\cite{Chauvin2019AnalyzingManifold,toews2007statistical}, consistent with uncertainty due to Gaussian blur. The localization accuracy of SIFT correspondences has been compared to that of manual human labeling in 2D and 3D ~\cite{toews2007statistical,Machado2018Non-rigidMatching}, showing similar accuracy, where experts preferred automatic SIFT correspondences in 80\% cases~\cite{Machado2018Non-rigidMatching} in the case of brain imaging. Our 3D SIFT-CPD algorithm proposed in the following section accounts for feature properties of orientation and scale, in addition to point locations.

Multi-modal image registration is a major challenge, where intensities may vary locally in a non-linear manner between modalities. Training may be used to approximate an intensity mapping for a specific domains, e.g. MRI and CT~\cite{hu2021end}, however this mapping may not be functional or stationary throughout the image. In the general case with no specific training domain or data, intensity-based registration must adopt statistical similarity measures such as mutual information~\cite{Viola1997AlignmentInformation}. In the case of 2D image keypoints, invariance to local contrast inversion may be achieved by transforming the image intensity into a contrast-invariant format, including phase congruency~\cite{xia2013robust} or Laplacian~\cite{Wachinger2012EntropyRegistration} images. As contrast inversion leads to image gradient reversal, attempts have been made to reverse aspects linked to the gradient, including descriptor orientation 
~\cite{kelman2007keypoint,chen2009real,hossain2012effective,Bingjian2011ImageImages}. Combinations of multiple keypoints~\cite{Teng2015MultimodalDescriptors} including self-similarity~\cite{Lv2019} may be adopted. In preliminary work, we proposed to account for 3D gradient reversal~\cite{toews2013feature}, here we extend this to include reflections and SP-symmetry via the notion of a binary sign $s \in \{-1,+1\}$.



\subsection{Discrete Symmetry}

In a physical system, a symmetry refers to a property that remains unchanged under set of transforms, i.e. an invariant property. The symmetry group is the Lie group of transforms under which a geometrical object such as a particle is invariant~\cite{liu2010computational}. A symmetry may be described as continuous or discrete. Continuous symmetry pertains to continuous parameters of pose and scale, eg. translation and orientation relative to the reference frame of an image acquisition device, and are analogous to a consequence of Noether's theorem for Lagrangian mechanical systems~\cite{noether1971invariant,tanaka2021noether}. Continuous symmetry has been investigated in the computer vision literature in terms of differential invariance~\cite{florack1992scale,olver1994differential,Lindeberg1998}. Classical neural networks such as CNNs are generally invariant to translations but not to reflections or scale changes~\cite{lecun1989backpropagation,krizhevsky2012imagenet}. Robustness to certain transforms may be improved via regularization~\cite{bardes2021vicreg}. Recently models of invariance or equivariance in trainable neural networks have emerged, including SO(3)~\cite{esteves2018learning,spezialetti2019learning} rotations, 3D rigid transforms\cite{moyer2021equivariant} non-linear transforms~\cite{spezialetti2019learning}, probabilistic symmetries~\cite{bloem2020probabilistic} and gauge theory~\cite{cohen2019gauge}. Much of this work has been limited to 2D image modalities, i.e. surfaces or projective images, and has not considered the 3D notion of SP-symmetry including a binary sign $s \in \{-1,+1\}$.

A discrete symmetry generally involves discrete displacements in location and/or orientation, e.g. translations, rotations, reflections, glide reflections or repetitive crystal structure. Charge conjugation and parity symmetry (CP-symmetry)~\cite{sozzi2008discrete}, and generally space-time reversal in CPT-symmetry including time relating to the transition of a particle to its anti-particle~\cite{borchert202216,Lehnert2022} are important aspects of discrete symmetry relating to 3D reflection of elementary physical particles. CP-symmetry is closely related to reflection symmetry, which exists in a function exhibiting one or more planes of symmetry, i.e. a  reflection about a single axis $x=0$ such that $f(x)=f(-x)$, or in a 3D system, the parity coordinate transform $f(x,y,z)=f(-x,-y,-z)$, both characterized in 3D by a rotation matrix $R$ with determinant $det(R)=-1$. The effect of the parity transform may be seen by first representing $f(x)$ in terms of $f(x) = f_s(x) + f_a(x)$, the sum purely symmetric and anti-symmetric image components $f_s(x)$ and $f_a(x)$ defined by $f_s(-x)=f_s(x)$ and $f_a(-x)=-f_a(x)$. Symmetric or even functions $f_s(x)$ include the Gaussian and Laplacian-of-Gaussian operators, wave functions associated with bosonic particles such as the photon, the cosine function, etc. Anti-symmetric or odd functions $f_a(x)$ include the gradient operator $\nabla I$, wave functions associated with fermionic particles such as electrons, and the sine function, etc. As the gradient is an anti-symmetric operator, an image contrast inversion $-\nabla I(x)=\nabla I(-x)$ is equivalent to an axis reflection. This phenomenon is analogous to charge conjugation and parity (CP) symmetry, and describes the transition of a charge-associated particle to its antiparticle, e.g. electrons and positrons. Notably, CP-symmetry was thought to be an inviolable in nature until the discovery of special cases beginning with the composite meson particle~\cite{christenson1964evidence}. In our work here, charge is equivalent to image sign or contrast, which may be inverted, and we propose SP-invariance to register images of differing modality.

Image processing methods specifically relating to our work have analyzed discrete shape symmetry (ie. reflections)~\cite{liu2010computational} and image contrast inversion~\cite{chen2009real,hossain2012effective,Teng2015MultimodalDescriptors,Alexander2001,Lv2019} as separate, unrelated phenomena. Our work is the first to consider these within the unified framework of discrete SP-symmetry, including a binary sign and discrete states of local reference frame orientation. Our method generalizes to various contexts where orientation is estimated from dominant local image gradients $\nabla I$. For example, Toews et al. identify maxima in a 3D spherical gradient histogram~\cite{Toews2013a} such that $\vect{\hat{\theta}}_1 =  \underset{\vect{\hat{\theta}}}{\mathrm{argmax}}~|\nabla I\cdot \hat{\theta}|$,  $\vect{\hat{\theta}}_2 =  \underset{\vect{\hat{\theta}}}{\mathrm{argmax}}~|\vect{\hat{\theta}} \times (\nabla I \times \vect{\hat{\theta}}_1)|$, with a third axis defined by the cross product $\vect{\hat{\theta}}_3 = \vect{\hat{\theta}}_1 \times \vect{\hat{\theta}}_2$. Rister et al. compute the eigenvectors of the local gradient structure tensor, ie. the $3\times3$ gradient correlation matrix~\cite{rister2017volumetric}. We demonstrate that both such representations may be modeled in terms of discrete SP-symmetry.


\section{Method}

Our method provides a means of characterizing 3D image features in a manner invariant to discrete symmetry transforms, including symmetric image patterns and sign inversion and parity (SP-symmetry) transforms due to image contrast inversion. We begin by describing the properties of a single local image feature, which we augment to include a binary sign and a set of discrete axis reflections. This allows us to compute a descriptor that is invariant to reflections and changes in sign due inversion of intensity contrast. We then define a kernel function $\mathcal{K}(f_n,f_m)$ to quantify the similarity of a pair of features $(f_n,f_m)$ potentially arising from the same underlying structure in two different images. Finally, this kernel is  incorporated into probabilistic image registration in order to estimate a transform $T: f_m \rightarrow f_n$ aligning the coordinate systems of two images.



\subsection{Single-Feature Properties}
\label{sec:particle}
A feature is a distinctive spherical region localized in 3D image space, as illustrated in Figure~\ref{fig:properties}. An individual feature $f=\{g,\vect{a},s\}$ is characterized within the image by its geometry $g$, a descriptor of the image appearance $\vect{a}$ surrounding the feature and a binary sign $s$. The feature geometry $g=\{\Theta,\sigma,\vect{x}\} \in SO(3)\times R^+ \times R^3$ is a scaled coordinate reference frame in 3D defined by 7 parameters. These include a scale $\sigma \in \mathbb{R}^+$, a coordinate location $\vect{x} \in \mathbb{R}^3$ and a reference frame $\Theta \in SO(3)$. The reference frame may be defined as a set $\Theta = \{\vect{\hat{\theta}}_1,\vect{\hat{\theta}}_2,\vect{\hat{\theta}}_3\}$ of orthonormal unit vectors $\vect{\hat{\theta}}_i, i \in [1,2,3]$ where $\vect{\hat{\theta}}_i \cdot \vect{\hat{\theta}}_j = 0, i \ne j$. Note equivalently 3D rotation could make use of alternative representations, eg. $SU(2)$ Pauli matrices or unit quaternions. Feature appearance $\vect{a}\in \mathbb{R}^M$ is an M-dimensional descriptor of the image, resampled according to geometry $g$. The sign $s \in \{-1,1\}$ is used to achieve invariance to image contrast inversion. 

Let $I: \mathbb{R}^3 \rightarrow \mathbb{R}^1$ be a scalar image sampled over 3D coordinates where $I(\vect{x})$ represents a voxel measurement at location $\vect{x}$. Let $I(\vect{x},\sigma)=I(\vect{x})*G(\vect{x},\sigma)$ be a scale-space defined by convolution of the image with a Gaussian filter $G(\vect{x},\sigma)$ of parameter $\sigma$~\cite{Lindeberg1998}. Features are detected as points in scale-space $\{\vect{x_m},\sigma_m,s_m\}$ at which the magnitude of the scale-normalized Laplacian operator is maximized, similarly to the original SIFT method~\cite{Lowe2004DistinctiveKeypoints}, here $m$ is a feature index. Furthermore, they are endowed with a binary sign as follows.
\begin{align}
    \{\vect{x_m},\sigma_m\} &=  \underset{\vect{x},\sigma}{\mathrm{ local~argmax}}~|\nabla^2 I(\vect{x},\sigma)|, \notag \\
    s_m &= {\mathrm{sign}} \left( \nabla^2 I(\vect{x}_m,\sigma_m) \right).
\end{align}
Scale-space loci $\{\vect{x_m},\sigma_m,s_m\}$ represent generic spherical regions in which the divergence of the local gradient field $\nabla I(\vect{x},\sigma)$ is maximized. Feature sign $s_m$ is novel to our characterization, and is defined as the sign of the Laplacian-of-Gaussian.


Feature orientation $\Theta$ is derived from the image gradient $\nabla I$ surrounding the feature origin $\vect{x}$. Orientation axis vectors $\vect{\hat{\theta}}_1,\vect{\hat{\theta}}_2,\vect{\hat{\theta}}_3$ may be determined according dominant gradient orientations within a window $w(\vect{x},\sigma) \in [0,1]$ centered upon $\vect{x}$ with spatial extent proportional to scale $\sigma$. As $\Theta \in SO(3)$, the determinant $det(\Theta)=1$ is positive, ie. no reflections. Axes are thus defined by two primary axis vectors $\vect{\hat{\theta}}_1$ and $\vect{\hat{\theta}}_2$, with the third constrained according to a handedness convention, ie. the right-hand rule via the cross product $\vect{\hat{\theta}}_3 = \vect{\hat{\theta}}_1 \times \vect{\hat{\theta}}_2$. Furthermore, we assume that axis vectors may be ordered uniquely according to the gradient magnitude along their directions, i.e. $[| \nabla I\cdot\vect{\hat{\theta}}_1|] > [| \nabla I\cdot\vect{\hat{\theta}}_2|]$, where $[ | \nabla I\cdot\vect{\hat{\theta}}_i|] = \int w(\vect{x},\sigma)| \nabla I(\vect{u})\cdot\vect{\hat{\theta}}_i| du$ is the expected value of the gradient magnitude along axis $\vect{\hat{\theta}}_i$.

\begin{figure*}
    \centering
    \includegraphics[width=0.9\textwidth]{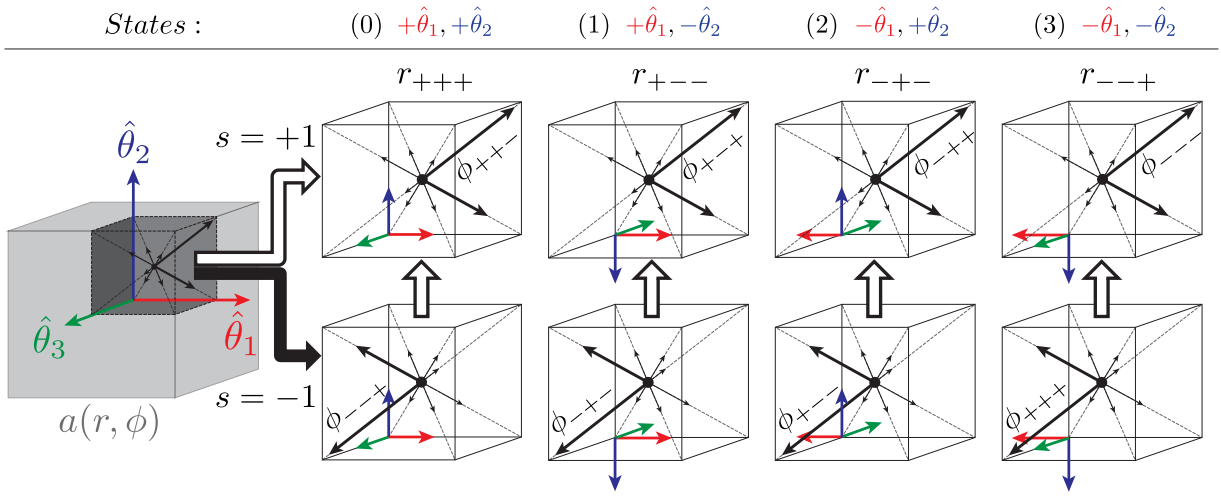}
    \caption{Illustrating an appearance descriptor $a(r,\phi)$ and four discrete orientation states given principal axes $\hat{\theta}_1$, $\hat{\theta}_2$ (left to right) and feature sign $s$ (upper and lower). The descriptor content is shown for an example octant ($r_{+++}$) and gradient orientation ($\phi_{++-}$). Axis orientation is defined independently of sign (left to right), and negative sign $s=-1$ inverts the descriptor gradient (lower row) achieving invariance to contrast inversion.
    }
    \label{fig:workflow}
\end{figure*}

Our primary contribution is a descriptor that is invariant to discrete SP-transforms, i.e. is SP-symmetric, based on binary sign $s$. We first note that the sign of the orientation axes $\vect{\hat{\theta}}_1,\vect{\hat{\theta}}_2$ may generally be ambiguous due to a variety of factors including bilateral symmetry of the image pattern or its local gradient distribution (e.g. an ellipsoid or rectangular box), an image contrast inversion (e.g. different imaging modalities), or the algorithm used to estimate orientation (e.g. principal component analysis has an inherent eigenvector sign ambiguity). Sign ambiguity may be represented by maintaining a discrete set of possible orientation reference frames $\Theta$. We consider four potential reference frame states $\{\pm \vect{\hat{\theta}_1},\pm \vect{\hat{\theta}_2}\}$ (with $\vect{\hat{\theta}_3}=\vect{\pm\hat{\theta}_1} \times \pm\vect{\hat{\theta}_2}$) resulting from axis inversion, i.e. due to either intensity contrast inversion or pattern symmetry, which are equivalent to the identity in addition to rotations of $\pi$ about each of three axes. By defining the third axis vector $\vect{\hat{\theta}_3}$ using the cross product, we ensure feature orientation follows the right-hand rule. Figure~\ref{fig:workflow} illustrates the four possible orientation states, including an initial reference frame and rotations of $\pi$ about each of the 3 axes.

Feature sign $s$ is used to generate a novel contrast inversion-invariant descriptor $\vect{a}_m$ for each orientation state, based on the image content surrounding location $\vect{x_m}$ at scale $\sigma_m$ and orientation $\Theta_m$. Note that the geometry $g$ defines a 7-parameter similarity transform from the local feature reference frame to a canonical reference frame, including rotation, translation and scaling. This this is used to resample the image according to a characteristic reference frame $\hat{I}(\vect{u}) = I(\sigma_m\Theta_m\vect{u}+\vect{x_m})$ that is invariant to global similarity transforms of the image. A variety of descriptors may be computed, eg. spherical harmonics~\cite{dimaio2009spherical}, we adopt a computationally efficient variant of the gradient orientation histogram~\cite{Lowe2004DistinctiveKeypoints,Toews2013a,rister2017volumetric}. As in~\cite{Toews2013a}, 3D space surrounding the feature origin $\vect{x}_m$ is quantized uniformly into eight discrete octants $\vect{r}=[\pm 1,\pm 1,\pm 1]$, and the image gradient at each point $\vect{u}$ into eight discrete symmetric directions $\vect{\phi}=[\pm 1,\pm 1,\pm 1]$. These are combined into an $8\times8=64$-element appearance descriptor, by accumulating the gradient magnitude into a histogram over spatial location and orientation
\begin{align}
a(\vect{r},\vect{\phi}) = \sum_{\vect{u} \in \vect{r}} \| \nabla \hat{I}(\vect{u})\cdot \vect{\phi}  \|[\vect{\phi}=s\vect{\phi}_{\vect{u}}],
\label{eq:descriptor}
\end{align}
where $\vect{\phi}_{\vect{u}}= \underset{\vect{\phi}}{\mathrm{argmax}}
\{ \nabla \hat{I}(\vect{u}) \cdot s\vect{\phi} \}$ is the direction of maximum gradient at image location $\vect{u}$, which is inverted in the case of $s=-1$ by the Iverson bracket $[\vect{\phi}=s\vect{\phi}_{\vect{u}}]$  evaluating to 1 upon equality and 0 otherwise. Thus descriptor orientation elements $\vect{\phi}$ are invariant to reversal due to sign change $a(\vect{r},\vect{\phi}) = a(\vect{r},s\vect{\phi})$, and parity transforms of orthants $\vect{r} \rightarrow -\vect{r}$ are accounted for by feature orientation states $\{\pm \vect{\hat{\theta}_1},\pm \vect{\hat{\theta}_2}\}$, as illustrated in Figure~\ref{fig:workflow}.



\subsection{Two-Feature Observations}

Image registration requires identifying pairs of features $(f_n,f_m)$ arising from the same underlying structure in different images. To do this, we propose a kernel function $K(f_n,f_m)\in [0,1]$ to quantify the similarity of two features $f_n$ and $f_m$ potentially sampled from the same distribution. We construct $K(f_n,f_m)$ as a product of squared exponential kernels, ensuring that it is positive, symmetric $K(f_m,f_n)=K(f_n,f_m)$, and proportional to a Gaussian density. It may be expressed as the product of three factors
\begin{align}
    \mathcal{K}(g_{n},g_{m}) = &\mathcal{K}_{\sigma}(\sigma_{n},\sigma_{m})\mathcal{K}_{\Theta}(\Theta_{n},\Theta_{m})\mathcal{K}_{x}(\vect{x_{n}},\vect{x_{m}}), \label{eq:geo_kernel}
\end{align}
quantifying the variability in scale, orientation and location, respectively. As in the notions of linear and angular momentum from classical physics, deviations of location and orientation are modeled as orthogonal and independent.

The first factor in Equation~\eqref{eq:geo_kernel},  $\mathcal{K}_{\sigma}(\sigma_{n},\sigma_{m})$, is a kernel penalizing the log scale difference:
\begin{equation}
   \mathcal{K}_{\sigma}(\sigma_{n},\sigma_{m}) = \exp \left( - \| \log \sigma_n - \log \sigma_m \|^2 \right),
   \label{eq:scale_kernel}
\end{equation}
and is proportional to a log normal density about mean $\log 1=0$ with unit variance $1$. The log maps multiplicative variations in feature scale on the range $\sigma \in [0,\infty]$ to additive variations on the range $\log \sigma \in [-\infty,\infty]$, which may be modelled as symmetric and Gaussian.

The second factor $\mathcal{K}_{\Theta}(\Theta_{n},\Theta_{m})$ quantifies the variability of feature orientations based on the angular displacement of the axes:
\begin{equation}
    \mathcal{K}_{\Theta}(\Theta_{n},\Theta_{m}) = \exp \left(-3 + \sum_{i=1,2,3} \vect{\hat{\theta}_{in}}\cdot \vect{\hat{\theta}_{im}} \right).
    \label{eq:orient_kernel}
\end{equation}
In Equation~\eqref{eq:orient_kernel}, the scalar product
$\vect{\hat{\theta}_{in}}\cdot \vect{\hat{\theta}_{im}}=\cos(\vect{\hat{\theta}_{in}}- \vect{\hat{\theta}_{im}})$ quantifies the angular separation between axis unit vectors $\vect{\hat{\theta}_{in}}$ and  $\vect{\hat{\theta}_{im}}$ on the range $[-1,1]$. This is equivalent to modeling the angular difference between features independently in each axis $i$ as proportional to a von Mises density over the unit circle.

The third factor $\mathcal{K}_{x}(\vect{x_{n}},\vect{x_{m}})$ is a kernel that quantifies the similarity of feature locations based on their squared displacement:
\begin{equation}
    \mathcal{K}_{x}(\vect{x_{n}},\vect{x_{m}}) = \exp \left( -\frac{||\vect{x_n} - \vect{x_m}||^2}{k~\sigma_{n}\sigma_{m}+\sigma_{T}^2} \right).
    \label{eq:loc_kernel}
\end{equation}
where the denominator normalizes the spatial displacement $||\vect{x_n} - \vect{x_m}||^2$ via a linear function of the product of feature scales $\sigma_{n}\sigma_{m}$.
The first term $k~\sigma_{n}\sigma_{m}$ embodies intrinsic variance due to the scale of the observed feature, where $\sigma_{n}\sigma_{m}$ represents the square of the geometric mean of feature scales and $k$ is a positive proportionality constant. This allows quantifying the deviation in location in a manner invariant to keypoint scale. The second term $\sigma_{T}^2$ represents the minimum achievable variance in mapping $T$ given the particular imaging context, which dominates in the case of small-scaled features, i.e. $k~\sigma_{n}\sigma_{m} < \sigma_{T}^2$. Note that in previous work~\cite{chauvin2021efficient} $k=1$ and $\sigma_{T}=0$, here we use $k$ and $\sigma_{T}$ to estimate a more accurate linear relationship between scale and observed variability. These are empirically set to provide a generally useful weighting, e.g. here we use $k=12$, $\sigma_T^2=200$.

\subsection{Multi-Feature Registration}

Registration seeks a transform $\mathcal{T}$ mapping the coordinate system of a fixed image $I_f$ to that of a moving image $I_m$, here from sets of features extracted in the fixed $\mathcal{F}=\{f_{n}\}$ and moving $\mathcal{M}=\{f_{m}\}$ images. The transform maps the properties of features from one coordinate space to the next $\mathcal{T}:\vect{x_m} \rightarrow \vect{x_n$}, and here is taken to be a global similarity transform, followed by independent zero-mean deviations for individual features. Note that the similarity transform is a 7-parameter representation  $\mathcal{T}=\{d\Theta,d\sigma,d\vect{x}\} \in SO(3)\times R^+ \times R^3$ equivalent that of an individual feature.

Let $g_m'$ represent the geometry of feature $f_m$ as transformed by $\mathcal{T}$, ie. $g_m'=T \circ g_m$, note that $T$ may act on location, scale and orientation parameters in a consistent manner. We propose using our kernel function $\mathcal{K}(g_n,g_m')$ to account for feature properties including scale and orientation in a standard point-based registration framework. Our kernel lends itself naturally to a probabilistic algorithm such as Coherent Point Drift (CPD) in Algorithm~\ref{alg:cpd}, which seeks a maximum likelihood solution based on the Expectation-Maximization (EM) algorithm~\cite{dempster1977maximum}. The expectation step (E) intend estimates the log-likelihood based on the current parameters, while the maximization step (M) maximizes the log-likelihood over the parameters using the result of the E step. CPD registration is based on a probability map $p_{mn}$ between each pair of points $\{f_n,f_m\}$, and takes the form of a softmax function. We propose to bias this probability map based on a kernel function $\mathcal{K}(g_n,g_m')$ as follows:
\begin{equation}
    p_{mn}=\frac{\exp\left(-\frac{||\vect{x_{\mathit{n}}}-\vect{x'_{\mathit{m}}}||^2}{2\lambda^2}\right)\mathcal{K}(g_{n},g'_{m})}{\sum_{k=1}^{M}\exp\left(-\frac{||\vect{x_{\mathit{n}}}-\vect{x'_{\mathit{k}}}||^2}{2\lambda^2}\right)\mathcal{K}(g_{n},g_{k}')+\eta},
    \label{eq:pij}
\end{equation}
where $\eta = (2\pi\lambda^2)^{D/2}\frac{w}{1-w}\frac{M}{N}$ is a constant ensuring the denominator $p_{mn}$ is non-zero, where $w$ is parameter accounting for the relative probability of a uniform background density not associated with any specific feature. In Equation~\eqref{eq:pij}, the exponential expression with parameter $\lambda$ is a variance parameter that is reduced iteratively to converge to a solution, and a constant kernel function $\mathcal{K}(g_n,g_m')=1$ is identical to the original CPD algorithm~\cite{myronenko2010point}. Thus in our proposed SIFT-CPD, two exponential factors operate on point coordinates $\vect{x}$ including Equation~\ref{eq:loc_kernel}.

Iterative algorithms such as EM must be initialized within a neighborhood of the solution in order to converge correctly. We initialize registration via a global 3D Hough transform between feature sets $\mathcal{F}$ and $\mathcal{M}$ in 3D~\cite{Toews2013a} analogously to the classic 2D SIFT algorithm~\cite{Lowe2004DistinctiveKeypoints}, where all features vote independently as to the predicted transform, after which a most likely transform is identified\footnote{The Hough transform method was originally proposed to track particles in bubble chamber photographs~\cite{hough1959machine}}. Here, each feature $f_{n} \in \mathcal{F}$ votes for a transform $\mathcal{T}_{nm}: g_n \rightarrow g_m$ mapping the geometry of feature $f_{n}$ to that of feature $f_{m}$, where $f_{m}=\underset{f}{\mathrm{argmin}}~\| \vect{a}_n-\vect{a} \|$ is the nearest neighbor of $f_{n}$ in terms of the Euclidean distance between appearance descriptors. Matches are identified between all features and discrete orientation states $\{\pm\hat{\theta}_1,\pm\hat{\theta}_2\}$ including parity transforms, and in a manner invariant to intensity contrast inversions due to the binary sign $s \in \{-1,+1\}$ in Equation~\eqref{eq:descriptor}. A dominant transform $\mathcal{T}^* \in \{\mathcal{T}_{nm}\}$ is then identified such that it is consistent with the largest number of matches, i.e. inlier correspondences. A pair of features $(f_{n},f_{m})$ represents an inlier of the transform $\mathcal{T}^*$ if their transform $\mathcal{T}_{nm}: g_n \rightarrow g_m$ differs by less than a threshold $\|\mathcal{T}_{nm}-\mathcal{T}^*\| < Thres$ as follows:
\begin{align}
    \mathcal{T}^* =
    \underset{T}{\mathrm{argmax}}~
    |\{T: \| \mathcal{T}_{nm}-\mathcal{T} \| < Thres\} |.
    \label{eq:thres}
\end{align}
The thresholding operation in Equation~\eqref{eq:thres} may be defined as the logical conjunction of thresholds independently applied in rotation, scaling and squared displacement
\begin{align}
  \| \mathcal{T}_{nm}-\mathcal{T} \| < Thres 
  \rightarrow &  \\
 &\cos(\vect{\hat{\theta}}_i-\vect{\hat{\theta}}_{inm}) < ~\epsilon_{cos\theta} \notag \\
& \wedge~|\log \sigma-\log\sigma_{nm}| < ~\epsilon_{log\sigma}  \notag \\
&\wedge~\|\vect{x}-\vect{x_{nm}}\|^2/\sigma\sigma_{nm} < \epsilon_{x/\sigma}, \notag
\end{align}
using thresholds $(\epsilon_{cos\theta},\epsilon_{log\sigma},\epsilon_{x/\sigma})$. These may be set generously to apply in a wide variety contexts, here we use
$(\epsilon_{cos\theta},\epsilon_{log\sigma},\epsilon_{x/\sigma}) = (0.7,log 1.5, 0.25)$. The Hough transform process may be implemented efficiently using hash tables and the mean-shift clustering algorithm~\cite{comaniciu2002mean}. Our modified CPD algorithm is provided in Algorithm~\ref{alg:cpd}, and the $solve$ functions for similarity transforms is found in Algorithm~\ref{alg:solve_rigid} in Appendix~\ref{app:solve}.

\begin{algorithm}
    \SetKwInOut{InHere}{Inputs}
    \SetKwInOut{OutHere}{Outputs}
    \SetKwInOut{Init}{Initialization}
    \SetKwInOut{Opti}{EM Optimization}
    \DontPrintSemicolon
    
    \InHere{ $\mathcal{F}=\{f_{n}\}$~\qquad\text{Fixed Features} \\  $\mathcal{M}=\{f_{m}\}$\qquad\text{Moving Features}
    }
    \OutHere{ $\mathcal{T}:\mathcal{M} \rightarrow \mathcal{F}$~\quad\text{Transform}
    }
    
    \Init{$0\leq w \leq 1$\;
    $\lambda^2 = \frac{1}{DNM}\sum_{n=1}^{N}\sum_{m=1}^{M}||\vect{x_n}-\vect{x_m}||^2$\;
    $\mathcal{T} \leftarrow \mathcal{T}^*$}
    
    \EMOpti{
    \EStep{
        $p_{mn}=\frac{\exp^{-\frac{1}{2\lambda^2}||\vect{x_{\mathit{n}}}-\vect{x'_{\mathit{m}}}||^2}\mathcal{K}(g_{n},g'_{m})}{\sum_{k=1}^{M}\exp^{-\frac{1}{2\lambda^2}||\vect{x_{\mathit{n}}}-\vect{x'_{\mathit{k}}}||^2}\mathcal{K}(g_{n},g'_{m})+\eta}$}
    \MStep{
        $\{\mathcal{T},\lambda^2\} = solve(\mathcal{F},\mathcal{M},P)$
    }
    }
    The aligned point set is $\mathcal{T}(\mathcal{M})$ 
    \\ The correspondence probability is $P$.
    
    \caption{Probabilistic SIFT-CPD registration algorithm, adapted from~\cite{myronenko2010point} to include kernel $\mathcal{K}(g_n,g'_m)$.}
    \label{alg:cpd}
\end{algorithm}
\vspace{\baselineskip}
\noindent


\section{Experiments}

We hypothesize that our kernel function $\mathcal{K}(g_n,g_m)$ incorporating feature scale and orientation will improve the accuracy of registration beyond point information alone, and that feature orientation state and binary sign will provide invariance to intensity contrast inversion and reflections, independently of the method used to identify feature orientation $\Theta$. Experiments validate these hypotheses in the context of 3D medical imaging data, which provide the opportunity to investigate features arising from realistic, diverse anatomical structure observed across subjects and 3D imaging modalities.

Three sets of experiments are performed. The first is based on synthetic similarity transforms, and establishes baseline accuracy of methods in the case of known ground truth. The second consists of inter-subject image registration trials, first between brain MRI and then between chest CT volumes of different people, with confounds including brain tumors and contrast variations due to multiple T1 and T2-weighted MRI modalities, demonstrating superior algorithm performance in a diverse set of contexts. Variability is estimated by repeating trials following synthetic transforms of images and multiple subjects. The third investigates orientation state changes in registration trials for different feature orientation estimation methods, i.e., principal components~\cite{rister2017volumetric} or maximum orientation directions~\cite{Toews2013a}.

Note that the true mapping $T$ between the anatomies of different subjects is generally non-linear and may not exist throughout the image, due to aspects of anatomy specific to individuals or in the case of occlusion or missing structure. As mentioned, features are assumed to follow a transform $T$ that is globally linear followed by random feature-specific deviations as specified by our kernel $\mathcal{K}(g_n,g_m)$. For inter-subject registration, ground truth is established from inlier correspondences identified via the Hough transform, which are visually validated for correctness and rejected or manually adjusted if necessary. This follows the work of~\cite{toews2007statistical,Machado2018Non-rigidMatching}, where automatic and manually labeled correspondences exhibit the same error range. The average Point Registration Error (PRE) measure is used to quantify registration performance, based on the sum of 3D point differences between the registration solution and ground truth.

In all experiments, feature sets are extracted using a GPU implementation of the SIFT-Rank algorithm for computational efficiency~\cite{carluer2021gpu}, and registered using five point-cloud registration methods: ICP with 20 and 100 iterations (respectively ICP20 and ICP100), the original CPD algorithm using feature centers $\vect{x}$, SIFT-CPD using full feature geometry and our kernel $\mathcal{K}(g_n,g_m)$, and SIFT-CPD* using only inlier features from the Hough transform. All algorithms are initialized to the Hough transform solution prior to iterative registration in order to ensure a fair comparison.

\subsection{Synthetic Image Registration}

A preliminary experiment was first performed establish the baseline performances of algorithms against known ground truth, here synthetic similarity transformations of a single image. A T1w brain image from the Human Connectome Project (HCP)~\cite{VanEssen2012} dataset was selected (see Table~\ref{tab:experiments} for more details), and 100 different synthetic transforms were applied, where each was generated by a random rotation about each axis on an angle range of $\rho \in \pm [10^{\circ},30^{\circ}]$ followed by a translation $\delta \in \pm[0~mm,10~mm]$. The registration error in rotation and translation was evaluated relative to the known transform applied, and the results are presented in Table~\ref{tab:samesubject_registration}. As expected, the lowest error is achieved for SIFT-CPD* (inliers alone), validating this method as a baseline error. SIFT-CPD and CPD perform similarly, with slightly less error for SIFT-CPD. ICP results are generally poor, failing to converge in 8 and 21 cases for 20 and 100 iterations.


\begin{table*}
\large
\begin{center}
\resizebox{2\columnwidth}{!}{
\bgroup
\def\arraystretch{1.8}
\begin{tabular}{|cc|ccc|ccc|cc|cc}
    \hline
    \multicolumn{2}{|P{3.5cm}|}{\multirow{2}{*}{\backslashbox[3.9cm]{Algorithm}{Error}}}&\multicolumn{3}{|c|}{\textit{Rotation Error (deg)}}&\multicolumn{3}{|c|}{\textit{Translation Error (mm)}}&\multicolumn{2}{|c|}{\textit{PRE (mm)}}&\multicolumn{2}{|c|}{\textit{SSD ($e^{11}$)}}\\ 
    \cline{3-12}
    \multicolumn{2}{|P{3.5cm}|}{}&\multicolumn{1}{|c|}{X-Axis}&\multicolumn{1}{|c|}{Y-Axis}&\multicolumn{1}{c|}{Z-Axis}&\multicolumn{1}{|c|}{X-Axis}&\multicolumn{1}{|c|}{Y-Axis}&\multicolumn{1}{|c|}{Z-Axis}&\multicolumn{2}{|c|}{Mean $\pm$ Std.Dev.}&\multicolumn{2}{|c|}{Mean $\pm$ Std.Dev.}\\
    \thickhline
    \multicolumn{2}{|P{3.5cm}|}{ICP20} & \multicolumn{1}{|c|}{$25.7 \pm 22.0$} & \multicolumn{1}{|c|}{$16.8 \pm 14.8$} & \multicolumn{1}{|c|}{$24.8 \pm 17.2$} & \multicolumn{1}{|c|}{$14.2 \pm 19.6$} & \multicolumn{1}{|c|}{$12.8 \pm 29.1$} & \multicolumn{1}{|c|}{$17.4 \pm 29.0$} & \multicolumn{2}{|c|}{$16.37 \pm 8.41$} &
    \multicolumn{2}{|c|}{$7.62 \pm 1.03$}\\
    \hline
    \multicolumn{2}{|P{3.5cm}|}{ICP100} & \multicolumn{1}{|c|}{$8.1 \pm 18.9$} & \multicolumn{1}{|c|}{$6.3 \pm 18.8$} & \multicolumn{1}{|c|}{$8.5 \pm 18.5$} & \multicolumn{1}{|c|}{$9.0 \pm 24.0$} & \multicolumn{1}{|c|}{$9.2 \pm 32.8$} & \multicolumn{1}{|c|}{$8.5 \pm 23.2$} & \multicolumn{2}{|c|}{$8.16 \pm 17.15$} &
    \multicolumn{2}{|c|}{$4.89 \pm 2.31$}\\
    \hline
    \multicolumn{2}{|P{3.5cm}|}{CPD} & \multicolumn{1}{|c|}{$0.06 \pm 0.03$} & \multicolumn{1}{|c|}{$0.06 \pm 0.03$} & \multicolumn{1}{|c|}{$0.06 \pm 0.03$} & \multicolumn{1}{|c|}{$0.8 \pm 0.6$} & \multicolumn{1}{|c|}{$1.3 \pm 1.1$} & \multicolumn{1}{|c|}{$0.7 \pm 0.5$} & \multicolumn{2}{|c|}{$1.81 \pm 0.88$} &
    \multicolumn{2}{|c|}{$1.38 \pm 0.39$}\\
    \hline
    \multicolumn{2}{|P{3.5cm}|}{\textbf{SIFT-CPD}} & \multicolumn{1}{|c|}{$\bf{0.02 \pm 0.01}$} & \multicolumn{1}{|c|}{$\bf{0.01 \pm 0.01}$} & \multicolumn{1}{|c|}{$\bf{0.02 \pm 0.01}$} & \multicolumn{1}{|c|}{$\bf{0.9 \pm 0.5}$} & \multicolumn{1}{|c|}{$\bf{0.8 \pm 0.6}$} & \multicolumn{1}{|c|}{$\bf{0.5 \pm 0.4}$} & \multicolumn{2}{|c|}{$\bf{1.05 \pm 0.32}$} &
    \multicolumn{2}{|c|}{$\bf{1.43 \pm 0.33}$}\\
    \hline
    \multicolumn{2}{|P{3.5cm}|}{\textbf{SIFT-CPD*}} & \multicolumn{1}{|c|}{$\bf{0.02 \pm 0.01}$} & \multicolumn{1}{|c|}{$\bf{0.01 \pm 0.01}$} & \multicolumn{1}{|c|}{$\bf{0.02 \pm 0.01}$} & \multicolumn{1}{|c|}{$\bf{0.9 \pm 0.4}$} & \multicolumn{1}{|c|}{$\bf{0.9 \pm 0.4}$} & \multicolumn{1}{|c|}{$\bf{0.4 \pm 0.3}$} & \multicolumn{2}{|c|}{$\bf{0.68 \pm 0.15}$} &
    \multicolumn{2}{|c|}{$\bf{0.96 \pm 0.39}$}\\
    \hline
\end{tabular}
\egroup
}
\end{center}
\caption{Comparison of five registration algorithms based on 100 synthetic transforms of a single T1w MRI brain volume. Error is listed terms of individual rotation and translation axes, overall point registration error (PRE), and the sum of squared intensity differences (SSD) following registration.
}
\label{tab:samesubject_registration}
\end{table*}

\subsection{Inter-subject Registration}

Four sets of experiments quantified the accuracy of inter-subject registration, i.e. registration of images of different individuals. The goal was to evaluate the accuracy and speed of registration in a diverse set of contexts, including different aspects of anatomy (brain, chest), abnormalities (tumor) and modalities (T1w and T2w MRI, CT). Information regarding datasets is provided in Table~\ref{tab:experiments}. Note that minimum registration error is generally non-zero due to differences in anatomy, and the goal is to identify a robust transform minimizing the PRE for sets of inlier correspondences.

\begin{table*}
\begin{center}
\resizebox{1.5\columnwidth}{!}{
\bgroup
\def\arraystretch{1.8}
\begin{tabular}{|c|c|c|c|c|c|c|}
    \hline
    \textbf{Dataset} & \textbf{Anatomy} & \textbf{Modality} & \textbf{Resolution} & \textbf{Voxel Size (mm)}& \textbf{Features} & \textbf{Inliers}\\
    \thickhline
    \cite{VanEssen2012} & Brain & T1w MRI &$260\times260\times311$&$0.7\times0.7\times0.7$&$6000$&$350$\\
    \hline
    \cite{regan2011genetic} & Chest & CT &$280\times280\times235$&$1.2\times1.2\times1.2$&$3000$&$70$\\
    \hline
    \cite{Menze2015TheBRATS} & Brain Tumor & T1w MRI &$240\times240\times155$&$1.0\times1.0\times1.0$&$1200$&$70$\\
    \hline
    \cite{VanEssen2012} & Brain & T1w-T2w MRI &$260\times260\times311$&$0.7\times0.7\times0.7$&$5000$&$115$\\
    \hline
\end{tabular}
\egroup
}
\end{center}
\caption{Inter-subject registration experimental data.}
\label{tab:experiments}
\end{table*}

\subsubsection{T1w Brain Images}

\begin{figure}[!h]
    \centering
    \includegraphics[width=0.5\textwidth]{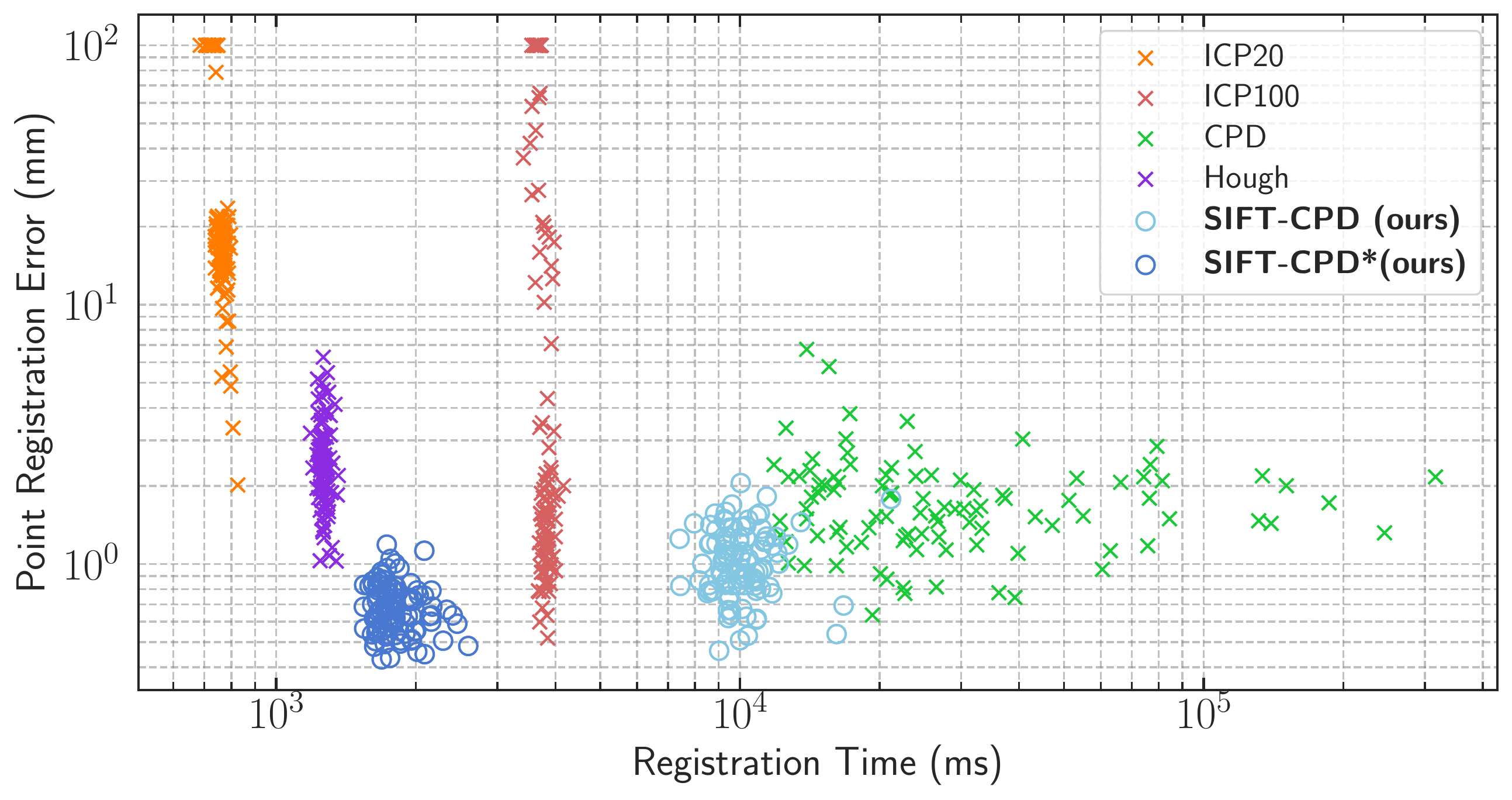}
    \caption{Inter-subject registration performance for healthy brain T1w MRI.
    }
    \label{fig:synth_data_brain}
\end{figure}

The following experiment quantified inter-subject registration accuracy between brain images of a pair of different subjects. As in the previous experiment, one image was fixed, while the other was subjected to a random transform from the previously mentioned ranges $\rho$ and $\delta$. A pair of monozygotic twin subjects was selected, which generally share more inlier correspondences in comparison to unrelated subjects due to higher neuroanatomical similarity~\cite{Chauvin2020NeuroimageRelatives}. The results for 100 random transforms are shown in Figure~\ref{fig:synth_data_brain}, these are generally consistent across all registration experiments. ICP variants which weight all points uniformly do not correctly converge (red and orange). CPD and SIFT-CPD (green and light blue) both converge consistently to acceptable results in term of PRE, both lower than that the initial robust Hough transform (purple). SIFT-CPD is consistently lower in terms of error and computation time in comparison to CPD, demonstrating that geometrical properties improve estimation, in the case where the majority of features may be outliers with no valid correspondence between images. Overall, the SIFT-CPD* (dark blue) leads to the lowest error and computation time, as it refines the original set of Hough transform inliers.

\subsubsection{CT Chest Images}
\begin{figure}
    \centering
    \includegraphics[width=0.5\textwidth]{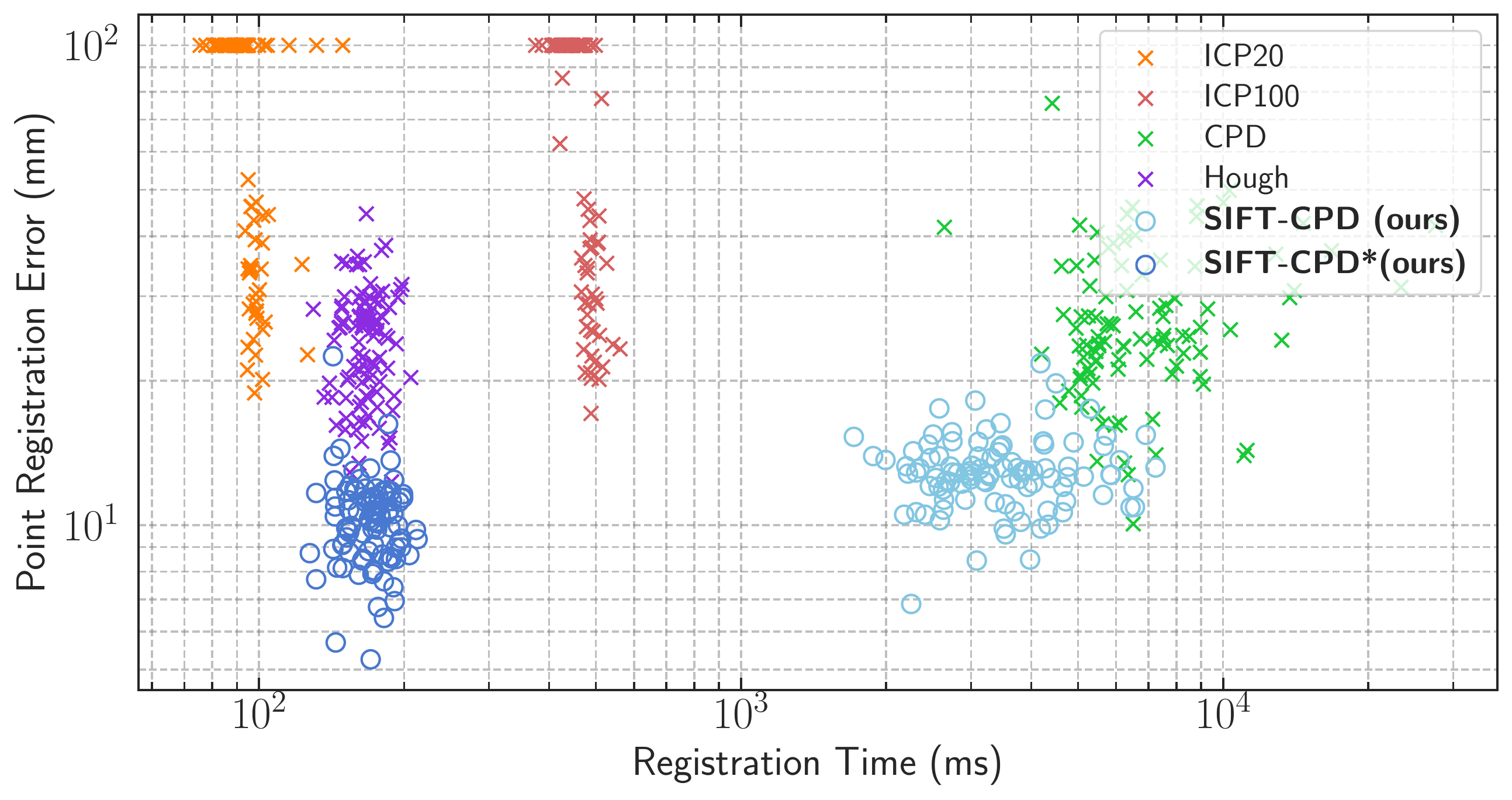}
    \caption{Inter-subject registration performance for chest CT images. 
    }
    \label{fig:synth_data_lungs}
\end{figure}

In order to show that our method is not tied to specific contexts or imaging modalities, we performed inter-subject registration of CT chest images from the Chronic Obstructive Pulmonary Disorder Genetics (COPDGene) dataset~\cite{regan2011genetic}. Two different subject images in expiration breathing state were randomly selected (see Table~\ref{tab:experiments} for more details).
As in the previous experiment, we generated 100 random transform with the same $\rho$ and $\delta$ parameters, applied the transform to one image, and registered it using different registration approaches. The PRE for each registration is shown on Figure~\ref{fig:synth_data_lungs}. Although the computational complexity of SIFT-CPD is comparable to the complexity of CPD in Figure~\ref{fig:synth_data_lungs}, SIFT-CPD* is significantly faster, closer to the ICP20 performances.


\subsubsection{Abnormal Variations and Occlusion}
\begin{figure}
    \centering
    \includegraphics[width=0.5\textwidth]{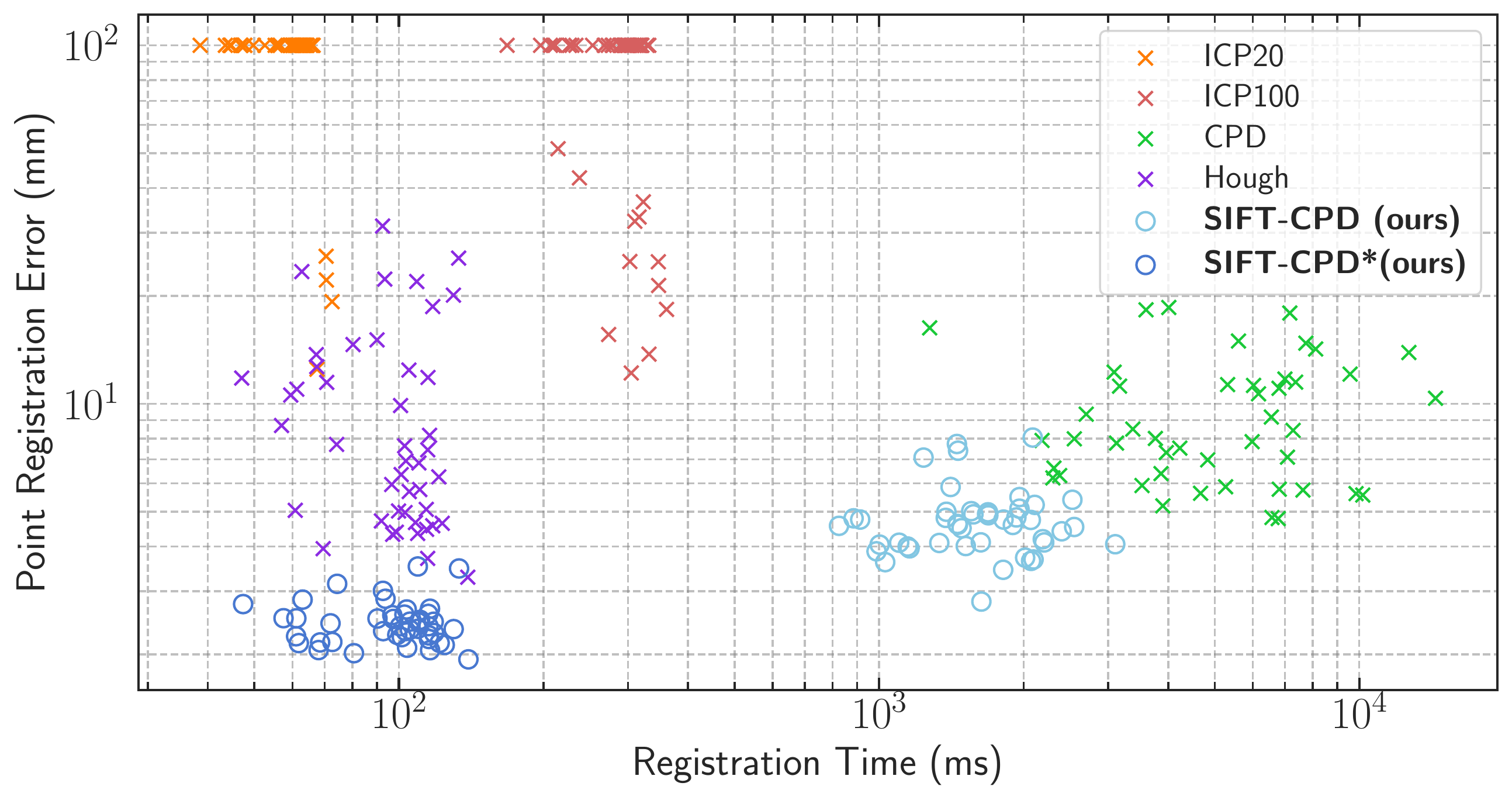}
    \caption{Inter-subject registration performance for brains in T1w MRI, with occlusions due to tumors~\cite{Menze2015TheBRATS}.}
    \label{fig:occlusion_results}
\end{figure}

Practical image analysis requires robustly coping with occlusions, including a lack of one-to-one correspondence due to inter-subject variations including pathological tumor structure. Our method is based on local feature properties, and is thus particularly robust to occlusion and missing structure. We used 10 T1w MRI images of different subjects with tumors from the BraTS dataset~\cite{Menze2015TheBRATS} in order to evaluate registration in the presence of occlusion (see Table~\ref{tab:experiments} for more details). Nine subject images are randomly selected, and transformed 5 times each, and registered to the remaining subject image. Results are presented in Figure~\ref{fig:occlusion_results}.


\subsubsection{Multi-Modal Images}
\begin{figure}[!h]
    \centering
    \includegraphics[width=0.5\textwidth]{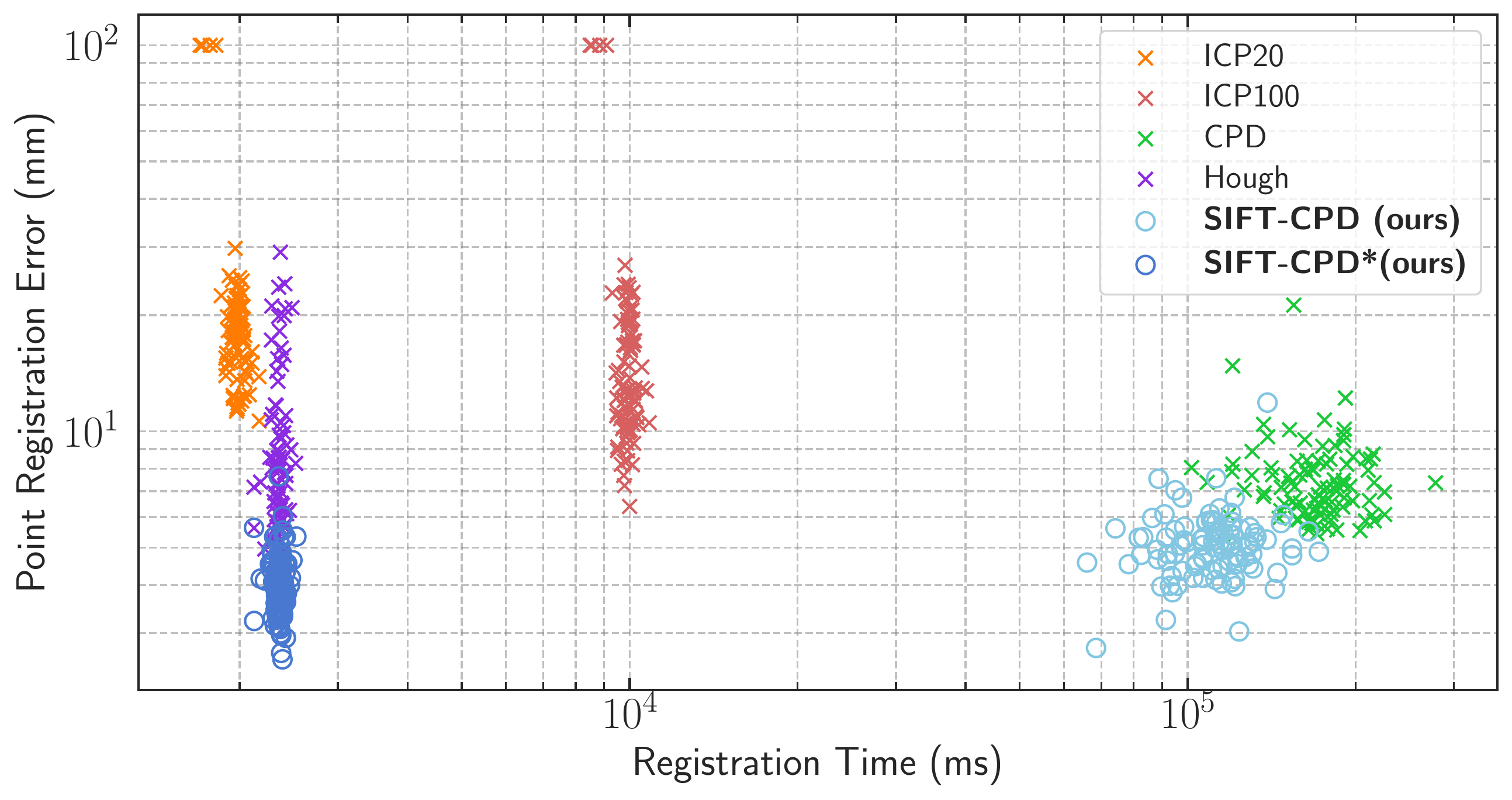}
    \caption{Inter-subject registration performance for healthy brain images between T1w and T2w MRI modalities~\cite{VanEssen2012}, exhibiting intensity contrast inversion.
    }
    \label{fig:synth_data_multimodal_brain}
\end{figure}

\begin{figure}[!h]
    \centering
    \includegraphics[width=0.5\textwidth]{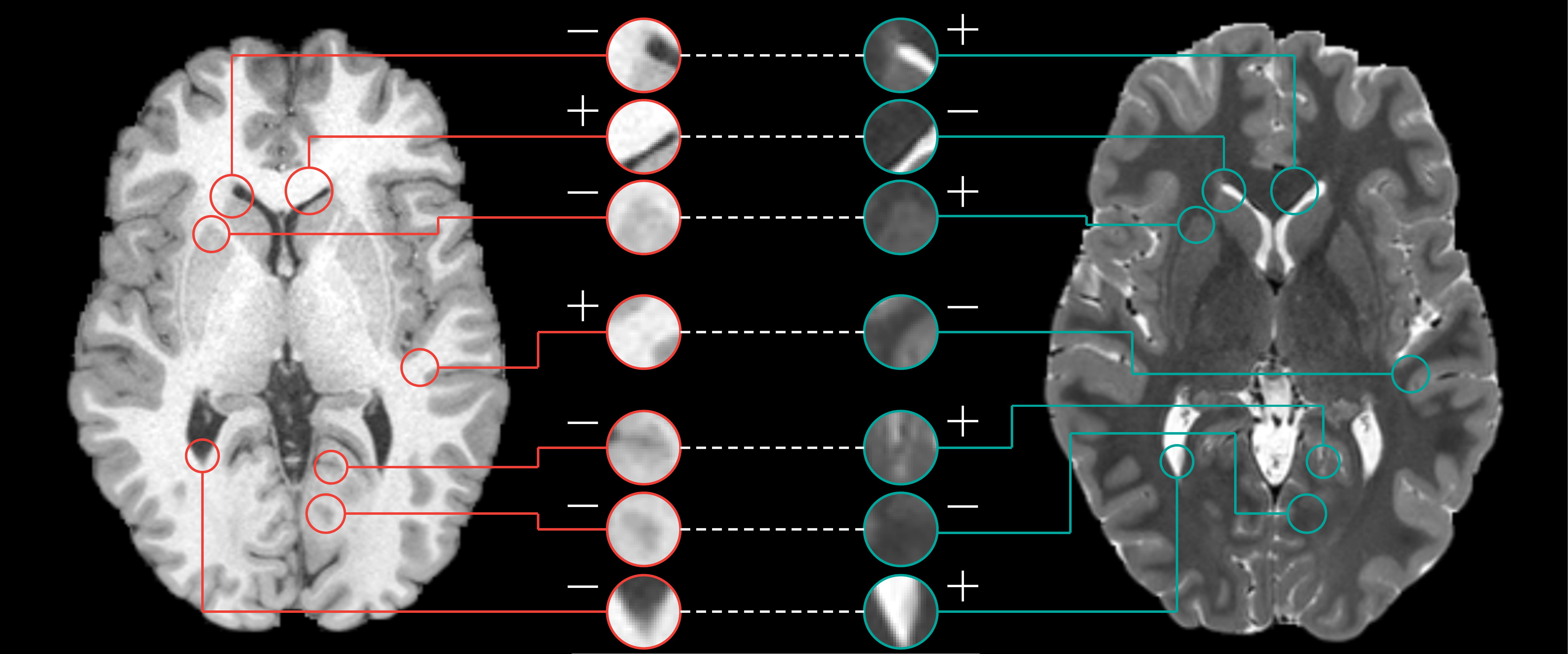}
    \caption{Examples of features matched between T1w and T2w images, showing feature signs, note the sign changes do not occur everywhere.}
    \label{fig:multimodal_matches}
\end{figure}

To evaluate the registration performance of our approach on multi-modal images, a random pair of brain images was selected from HCP dataset, one with a T1w image and the other one with a T2w (see Table~\ref{tab:experiments} for more details). T1w and T2w images are acquired with different MRI parameters and highlight different tissue properties, where T2w imaging involves longer repetition time (TR) and time-to-echo (TE) parameters. The T2w image was randomly transformed and registered as in previous experiments, with 100 trials. Results are presented in Figure~\ref{fig:synth_data_multimodal_brain}. An example of corresponding features identified in different modalities is shown in Figure~\ref{fig:multimodal_matches} and are investigated in greater detail in the following experiment section.

\subsection{Feature Sign and Orientation State}

Here we investigated the binary sign $s$ and orientation state $(\theta_1,\theta_2)$ of local features in registration experiments, using two different methods for estimating feature orientation $\Theta$: eigenvectors of the local gradient structure tensor matrix~\cite{rister2017volumetric} or maximum orthogonal gradient directions~\cite{Toews2013a}. Both methods have been demonstrated to achieve effective local feature correspondences between images of the same modality, but not different modalities where discrete orientation state changes may occur.

\begin{figure}
    \centering
    \includegraphics[width=0.5\textwidth]{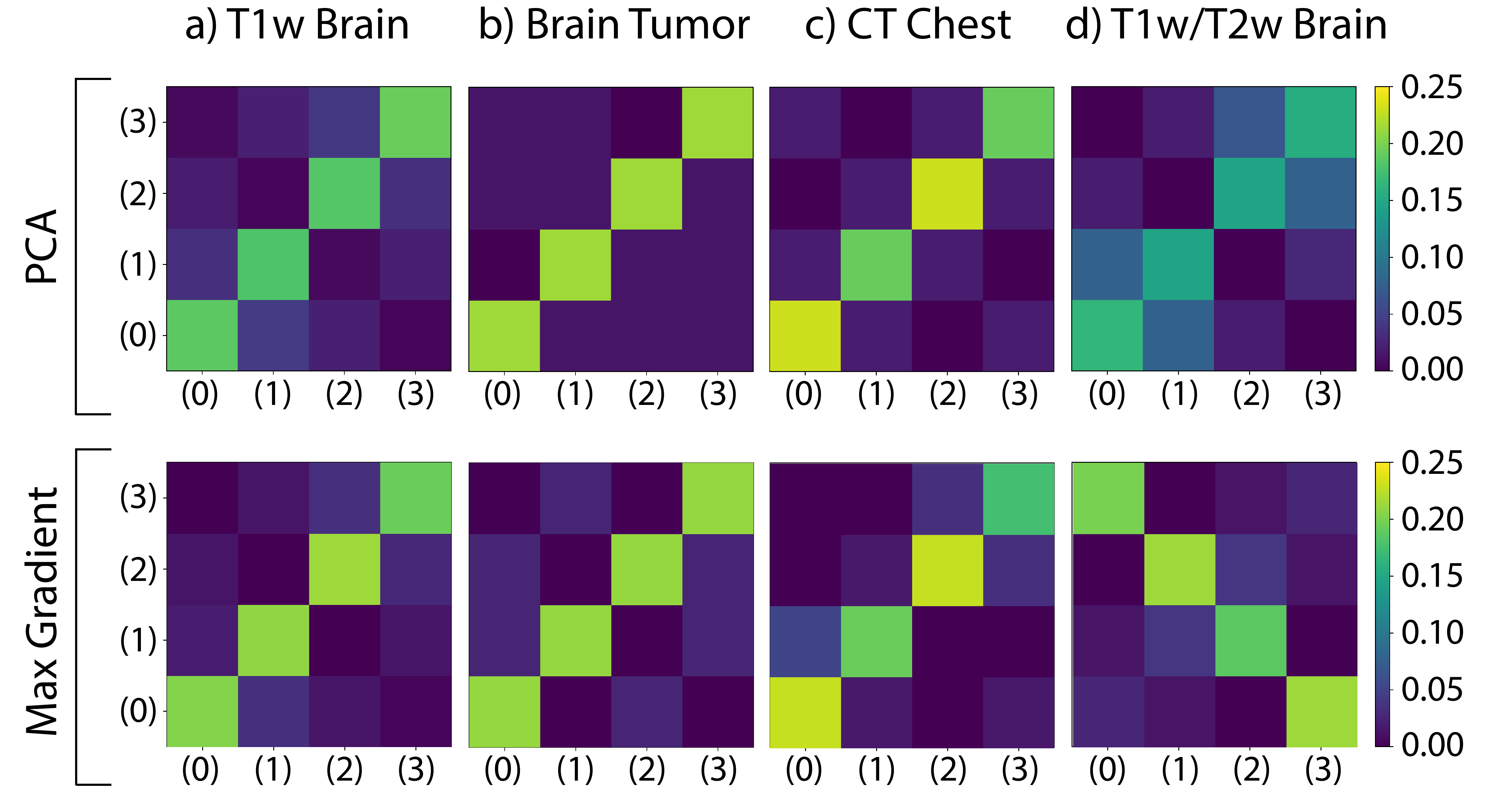}
    \caption{Orientation state transition histograms for principal gradient orientation estimation (see~\cite{rister2017volumetric}) and max gradient orientation estimation (see~\cite{Toews2013a}) for: a) T1w brain, b) T1w brain tumor~\cite{Menze2015TheBRATS}, c) CT chest~\cite{regan2011genetic}, d) Multi-modal T1w/T2w brain. States
    described in Figure~\ref{fig:workflow}, are combinations of discrete axis reflections $\{\pm\hat{\theta}_1,\pm\hat{\theta}_2\}$ as 0:$\{++\}$,
    1:$\{+-\}$,
    2:$\{-+\}$,
    3:$\{--\}$. Bins along the diagonal indicate no state change, off-diagonals indicate single-axis reflections and inverse diagonal indicates parity transform.
    }
    \label{fig:states}
\end{figure}

\begin{figure}
    \centering
    \includegraphics[width=0.5\textwidth]{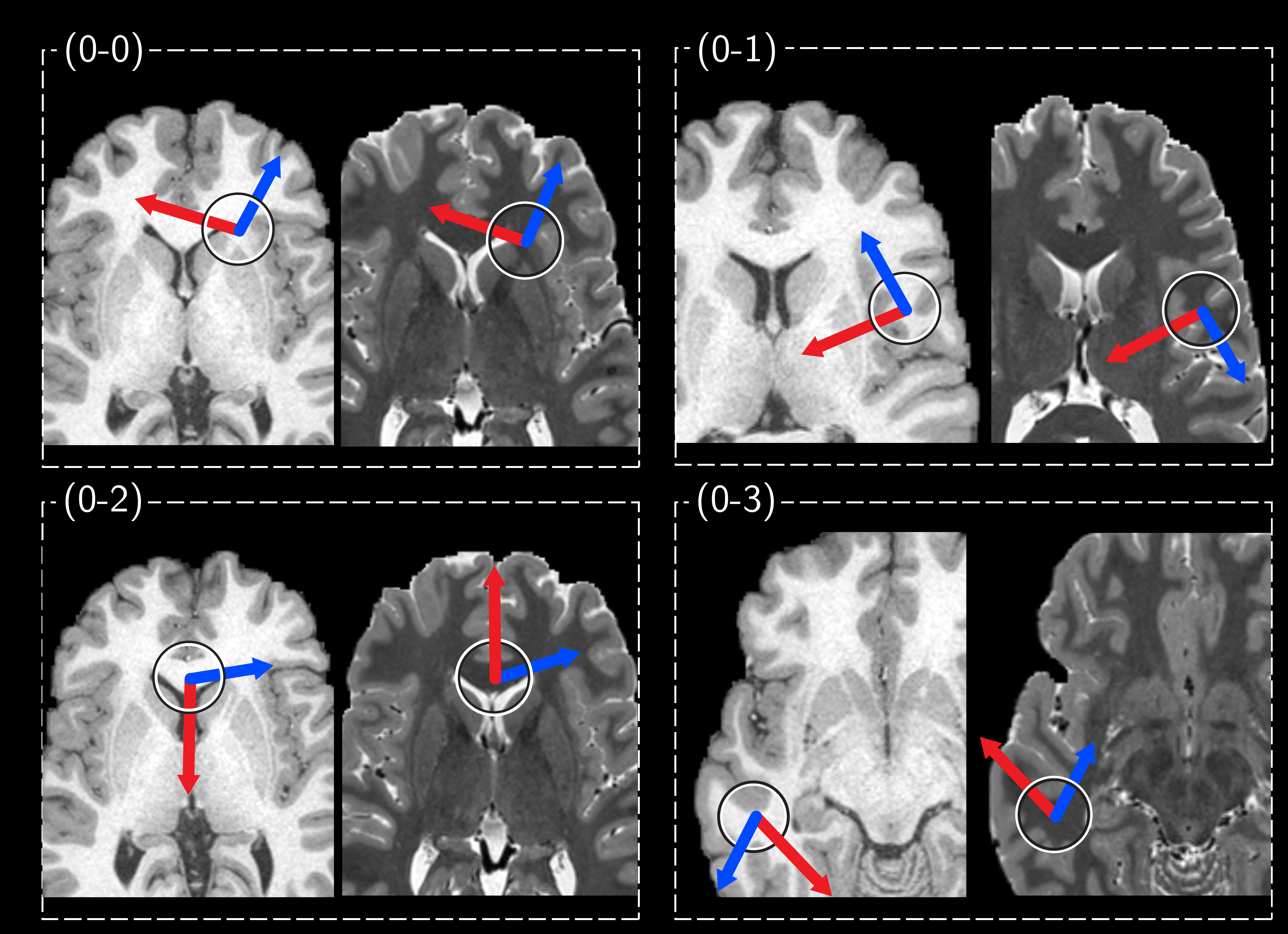}
    \caption{Examples of feature orientation state changes observed between T1w and T2w modalities. Red and blue arrows indicate the primary $\hat{\theta}_1$ and secondary $\hat{\theta}_2$ orientation axes projected onto an image slice plane. Circles indicate the feature location and scale in a slice plane, sign $s \in \{-1,+1\}$ is indicated by the white-black or black-white transitions. }
    \label{fig:symmetries}
\end{figure}

Figures~\ref{fig:states} show distributions of orientation state transitions of inlier correspondences between images. For same modality registration (a, b, c) there are relatively few state changes (main diagonal), however in the case of multi-modal registration (d) noticeable differences exist. In the case of orientation determined via principal component analysis (PCA) of gradients~\cite{rister2017volumetric}, approximately 40\% of all correspondences exhibit single-axis reflections (Figure~\ref{fig:states} d) rather than SP transforms, this is due to eigenvector sign ambiguity, which is accounted for in our algorithm. In the case of orientation determined by maximum gradient directions, correspondences generally exhibit full SP-transforms, i.e. sign inversion and parity (Figure~\ref{fig:states} d). Figure~\ref{fig:symmetries} shows examples of corresponding features and state transitions in the case of sign inversion, including identity (0-0), single axis reflections of major (0-1) and minor axes (0-2), and parity transforms (0-3).

\section{Discussion}

This paper proposed to extend local feature methods to account for SP-symmetry and to achieved 3D multi-modal image registration. We introduced the notion of discrete SP-symmetry in the of case features extracted from a scalar intensity image $I(\vect{x}) \in R^1$ over $\vect{x} \in R^3$ space. We proposed to model the geometrical properties of 3D image features in a manner invariant to SP-transforms observed in multi-modal image registration. We proposed to include a feature sign $s \in \{-,+\} = {\text sign}~\nabla^2~I(\vect{x},\sigma)$ analogous to electric charge, and maintain a discrete set of four orientation states including reflections of primary axes and parity transforms. Sign is used to invert gradient descriptor elements in the case of intensity contrast inversion, i.e. charge conjugation. Feature geometry and appearance descriptors are thus invariant to SP transforms, in addition to continuous rotations, scalings and translations as in the 3D SIFT method~\cite{Toews2013a,rister2017volumetric}.


We integrated feature properties into a well-known probabilistic point-based image registration framework, the CPD algorithm~\cite{myronenko2010point}, via a kernel function $\mathcal{K}(f_n,f_m)$, leading to a highly robust and stable SIFT-CPD registration algorithm. Image registration experiments are performed using a range of volumetric image modalities of the human brain and chest, showing that additional properties consistently improve the accuracy of registration in comparison to point locations alone. Registration error and computation time is consistently lower for our proposed SIFT-CPD vs. standard point-based CPD frameworks, indicating that feature scale and orientation improve the estimation of image-to-image transforms. An optimized version of SIFT-CPD functioning solely on sparse inlier feature correspondences exhibits the lowest error, refining the solution achieved via a robust but coarse initial Hough transform. Experiments also demonstrated how SP-symmetry may be applied despite the method used to estimate feature orientation estimation, here principal gradient orientations~\cite{rister2017volumetric} and maximum gradient orientations~\cite{Toews2013a}.  Invariant feature appearance descriptors are used to identify inliers and to initialize registration, there were not explicitly incorporated into the SIFT-CPD kernel function based solely on geometrical properties here, however this possibility is left for future work.




We note that the theory of SP-symmetry we present applies generally to individual channels of convolutional neural networks, and could be used within the general neural network learning framework to achieve SP-invariance, in addition to continuous invariance~\cite{bardes2021vicreg,esteves2018learning,spezialetti2019learning,cohen2019gauge}. Here, 3D invariant features have been identified via unbiased, symmetric operators with no explicit training procedure, which could be used as is in general imaging contexts, or to train or validate domain-specific detectors via neural networks~\cite{Yi2016LIFT:Transform,Detone2018SuperPoint:Description}.\\~\\
All code required to reproduce our results may be obtained at \href{https://github.com/3dsift-rank/SIFT-CPD}{https://github.com/3dsift-rank/SIFT-CPD}.

\ifCLASSOPTIONcompsoc
  \section*{Acknowledgments}
\else
  \section*{Acknowledgment}
\fi
This work was supported by NIH grant P41EB015902 (NAC), the Quebec (FRQNT) New Researchers Startup Program and the Canadian National Sciences and Research Council (NSERC) Discovery Grant. Data were provided by the Human Connectome Project, WU-Minn Consortium (Principal Investigators: David Van Essen and Kamil Ugurbil; 1U54MH091657) funded by the 16 NIH Institutes and Centers that support the NIH Blueprint for Neuroscience Research; and by the McDonnell Center for Systems Neuroscience at Washington University.

\ifCLASSOPTIONcaptionsoff
  \newpage
\fi



\bibliographystyle{IEEEtran}
\bibliography{mendeley,cpd}

\begin{thebibliography}{10}
\providecommand{\url}[1]{#1}
\csname url@samestyle\endcsname
\providecommand{\newblock}{\relax}
\providecommand{\bibinfo}[2]{#2}
\providecommand{\BIBentrySTDinterwordspacing}{\spaceskip=0pt\relax}
\providecommand{\BIBentryALTinterwordstretchfactor}{4}
\providecommand{\BIBentryALTinterwordspacing}{\spaceskip=\fontdimen2\font plus
\BIBentryALTinterwordstretchfactor\fontdimen3\font minus
  \fontdimen4\font\relax}
\providecommand{\BIBforeignlanguage}[2]{{%
\expandafter\ifx\csname l@#1\endcsname\relax
\typeout{** WARNING: IEEEtran.bst: No hyphenation pattern has been}%
\typeout{** loaded for the language `#1'. Using the pattern for}%
\typeout{** the default language instead.}%
\else
\language=\csname l@#1\endcsname
\fi
#2}}
\providecommand{\BIBdecl}{\relax}
\BIBdecl

\bibitem{schrodinger1926undulatory}
E.~Schr{\"o}dinger, ``{An undulatory theory of the mechanics of atoms and
  molecules},'' \emph{Physical review}, vol.~28, no.~6, p. 1049, 1926.

\bibitem{Toews2013a}
M.~Toews and W.~Wells, ``{Efficient and robust model-to-image alignment using
  3D scale-invariant features},'' \emph{Medical Image Analysis (MIA)}, vol.~17,
  no.~3, pp. 271--282, 2013.

\bibitem{rister2017volumetric}
B.~Rister, M.~A. Horowitz, and D.~L. Rubin, ``{Volumetric image registration
  from invariant keypoints},'' \emph{IEEE Transactions on Image Processing},
  vol.~26, no.~10, pp. 4900--4910, 2017.

\bibitem{Lindeberg1998}
T.~Lindeberg, ``{Feature Detection with Automatic Scale Selection},''
  \emph{International Journal of Computer Vision (IJCV)}, vol.~30, no.~2, pp.
  79 -- 116, 1998.

\bibitem{Lowe2004DistinctiveKeypoints}
D.~G. Lowe, ``{Distinctive image features from scale-invariant keypoints},''
  \emph{International Journal of Computer Vision (IJCV)}, vol.~60, no.~2, pp.
  91--110, 2004.

\bibitem{flitton2010object}
G.~T. Flitton, T.~P. Breckon, and N.~M. Bouallagu, ``{Object Recognition using
  {3D} {SIFT} in Complex {CT} Volumes.}'' in \emph{BMVC}, 2010, pp.
  11.1--11.12.

\bibitem{Machado2018Non-rigidMatching}
I.~Machado \emph{et~al.}, ``{Non-rigid registration of 3D ultrasound for
  neurosurgery using automatic feature detection and matching},''
  \emph{International Journal of Computer Assisted Radiology and Surgery
  (IJCARS)}, vol.~13, no.~10, pp. 1525--1538, 2018.

\bibitem{luo2018feature}
J.~Luo \emph{et~al.}, ``{A feature-driven active framework for ultrasound-based
  brain shift compensation},'' in \emph{International Conference on Medical
  Image Computing and Computer-Assisted Intervention}.\hskip 1em plus 0.5em
  minus 0.4em\relax Springer, 2018, pp. 30--38.

\bibitem{wachinger2018keypoint}
C.~Wachinger, M.~Toews, G.~Langs, W.~Wells, and P.~Golland, ``{Keypoint
  transfer for fast whole-body segmentation},'' \emph{IEEE transactions on
  medical imaging}, vol.~39, no.~2, pp. 273--282, 2018.

\bibitem{gill2014robust}
G.~Gill, M.~Toews, and R.~R. Beichel, ``{Robust initialization of active shape
  models for lung segmentation in CT scans: A feature-based atlas approach},''
  \emph{International journal of biomedical imaging}, vol. 2014, 2014.

\bibitem{chauvin2021efficient}
L.~Chauvin, K.~Kumar, C.~Desrosiers, W.~Wells~III, and M.~Toews, ``{Efficient
  Pairwise Neuroimage Analysis using the Soft Jaccard Index and 3D Keypoint
  Sets},'' \emph{arXiv preprint arXiv:2103.06966}, 2021.

\bibitem{Toews2015}
M.~Toews, C.~Wachinger, R.~S.~J. Estepar, and W.~Wells, ``{A Feature-Based
  Approach to Big Data Analysis of Medical Images.}'' \emph{International
  Conference on Information Processing in Medical Imaging (IPMI)}, vol.~24, pp.
  339--350, 2015.

\bibitem{borchert202216}
M.~Borchert \emph{et~al.}, ``{A 16-parts-per-trillion measurement of the
  antiproton-to-proton charge--mass ratio},'' \emph{Nature}, vol. 601, no.
  7891, pp. 53--57, 2022.

\bibitem{Lehnert2022}
\BIBentryALTinterwordspacing
R.~Lehnert, ``{Mirror symmetry validated for proton and its antimatter twin},''
  \emph{Nature 2021 601:7891}, vol. 601, pp. 32--33, 1 2022.
\BIBentrySTDinterwordspacing

\bibitem{christenson1964evidence}
J.~H. Christenson, J.~W. Cronin, V.~L. Fitch, and R.~Turlay, ``{Evidence for
  the 2 $\pi$ Decay of the K 2 0 Meson},'' \emph{Physical Review Letters},
  vol.~13, no.~4, p. 138, 1964.

\bibitem{sozzi2008discrete}
M.~Sozzi, \emph{{Discrete symmetries and CP violation: From experiment to
  theory}}.\hskip 1em plus 0.5em minus 0.4em\relax Oxford University Press,
  2008.

\bibitem{liu2010computational}
Y.~Liu, H.~Hel-Or, C.~S. Kaplan, L.~Van~Gool \emph{et~al.}, ``{Computational
  symmetry in computer vision and computer graphics},'' \emph{Foundations and
  Trends{\textregistered} in Computer Graphics and Vision}, vol.~5, no. 1--2,
  pp. 1--195, 2010.

\bibitem{cohen2019gauge}
T.~Cohen, M.~Weiler, B.~Kicanaoglu, and M.~Welling, ``{Gauge equivariant
  convolutional networks and the icosahedral CNN},'' in \emph{International
  conference on Machine learning}.\hskip 1em plus 0.5em minus 0.4em\relax PMLR,
  2019, pp. 1321--1330.

\bibitem{chen2009real}
J.~Chen and J.~Tian, ``{Real-time multi-modal rigid registration based on a
  novel symmetric-SIFT descriptor},'' \emph{Progress in Natural Science},
  vol.~19, no.~5, pp. 643--651, 2009.

\bibitem{hossain2012effective}
M.~T. Hossain, ``{An effective technique for multi-modal image registration},''
  Ph.D. dissertation, Monash University, 2012.

\bibitem{Teng2015MultimodalDescriptors}
\BIBentryALTinterwordspacing
S.~W. Teng, M.~T. Hossain, and G.~Lu, ``{Multimodal image registration
  technique based on improved local feature descriptors},''
  \emph{https://doi.org/10.1117/1.JEI.24.1.013013}, vol.~24, no.~1, p. 013013,
  1 2015.
\BIBentrySTDinterwordspacing

\bibitem{Alexander2001}
\BIBentryALTinterwordspacing
A.~Alexander, K.~Hasan, and M.~Lazar, ``{Analysis of partial volume effects in
  diffusion‐tensor MRI},'' \emph{Magnetic}, 2001.
\BIBentrySTDinterwordspacing

\bibitem{Lv2019}
G.~Lv, ``{Self-similarity and symmetry with SIFT for multi-modal image
  registration},'' \emph{IEEE Access}, vol.~7, pp. 52\,202--52\,213, 2019.

\bibitem{myronenko2010point}
A.~Myronenko and X.~Song, ``{Point set registration: Coherent point drift},''
  \emph{IEEE transactions on pattern analysis and machine intelligence},
  vol.~32, no.~12, pp. 2262--2275, 2010.

\bibitem{harris1988combined}
C.~Harris, M.~Stephens \emph{et~al.}, ``{A combined corner and edge
  detector},'' in \emph{Alvey vision conference}, vol.~15, no.~50.\hskip 1em
  plus 0.5em minus 0.4em\relax Citeseer, 1988, pp. 10--5244.

\bibitem{shi1994good}
J.~Shi \emph{et~al.}, ``{Good features to track},'' in \emph{1994 Proceedings
  of IEEE conference on computer vision and pattern recognition}.\hskip 1em
  plus 0.5em minus 0.4em\relax IEEE, 1994, pp. 593--600.

\bibitem{Mikolajczyk2005}
\BIBentryALTinterwordspacing
K.~Mikolajczyk and C.~Schmid, ``{A performance evaluation of local
  descriptors},'' \emph{IEEE Transactions on Pattern Analysis and Machine
  Intelligence}, vol.~27, no.~10, pp. 1615--1630, 10 2005.
\BIBentrySTDinterwordspacing

\bibitem{alcantarilla2012kaze}
P.~F. Alcantarilla, A.~Bartoli, and A.~J. Davison, ``{KAZE features},'' in
  \emph{European conference on computer vision}.\hskip 1em plus 0.5em minus
  0.4em\relax Springer, 2012, pp. 214--227.

\bibitem{kadir2001saliency}
T.~Kadir and M.~Brady, ``{Saliency, scale and image description},''
  \emph{International Journal of Computer Vision}, vol.~45, no.~2, pp. 83--105,
  2001.

\bibitem{toews2010mutual}
M.~Toews and W.~M. Wells, ``{A mutual-information scale-space for image feature
  detection and feature-based classification of volumetric brain images},'' in
  \emph{2010 IEEE Computer Society Conference on Computer Vision and Pattern
  Recognition-Workshops}.\hskip 1em plus 0.5em minus 0.4em\relax IEEE, 2010,
  pp. 111--116.

\bibitem{Bay2008}
\BIBentryALTinterwordspacing
H.~Bay, A.~Ess, T.~Tuytelaars, and L.~V. Gool, ``{Speeded-up robust features
  (SURF)},'' \emph{Computer vision and image}, 2008.
\BIBentrySTDinterwordspacing

\bibitem{jiang2021cotr}
W.~Jiang, E.~Trulls, J.~Hosang, A.~Tagliasacchi, and K.~M. Yi, ``{Cotr:
  Correspondence transformer for matching across images},'' in
  \emph{Proceedings of the IEEE/CVF International Conference on Computer
  Vision}, 2021, pp. 6207--6217.

\bibitem{ono2018lf}
Y.~Ono, E.~Trulls, P.~Fua, and K.~M. Yi, ``{LF-Net: Learning local features
  from images},'' \emph{Advances in neural information processing systems},
  vol.~31, 2018.

\bibitem{noh2017large}
H.~Noh, A.~Araujo, J.~Sim, T.~Weyand, and B.~Han, ``{Large-scale image
  retrieval with attentive deep local features},'' in \emph{Proceedings of the
  IEEE international conference on computer vision}, 2017, pp. 3456--3465.

\bibitem{Yi2016LIFT:Transform}
K.~M. Yi, E.~Trulls, V.~Lepetit, and P.~Fua, ``{LIFT: Learned invariant feature
  transform},'' in \emph{Proceedings of the European Conference on Computer
  Vision (ECCV)}, vol. 9910 LNCS.\hskip 1em plus 0.5em minus 0.4em\relax
  Springer Verlag, 2016, pp. 467--483.

\bibitem{Detone2018SuperPoint:Description}
D.~Detone, T.~Malisiewicz, and A.~Rabinovich, ``{SuperPoint: Self-Supervised
  Interest Point Detection and Description},'' in \emph{Proceedings of the IEEE
  Conference on Computer Vision and Pattern Recognition (CVPR) Workshops},
  2018.

\bibitem{carluer2021gpu}
J.-B. Carluer \emph{et~al.}, ``{GPU optimization of the 3D Scale-invariant
  Feature Transform Algorithm and a Novel BRIEF-inspired 3D Fast Descriptor},''
  \emph{arXiv preprint arXiv:2112.10258}, 2021.

\bibitem{luo2016understanding}
W.~Luo, Y.~Li, R.~Urtasun, and R.~Zemel, ``{Understanding the effective
  receptive field in deep convolutional neural networks},'' \emph{Advances in
  neural information processing systems}, vol.~29, 2016.

\bibitem{calonder2011brief}
M.~Calonder, V.~Lepetit, M.~Ozuysal, T.~Trzcinski, C.~Strecha, and P.~Fua,
  ``{BRIEF: Computing a local binary descriptor very fast},'' \emph{IEEE
  transactions on pattern analysis and machine intelligence}, vol.~34, no.~7,
  pp. 1281--1298, 2011.

\bibitem{Toews2009}
M.~Toews and W.~Wells, ``{Sift-rank: Ordinal description for invariant feature
  correspondence},'' in \emph{Proceedings of the IEEE Conference on Computer
  Vision and Pattern Recognition (CVPR)}, 2009, pp. 172--177.

\bibitem{arandjelovic2012three}
R.~Arandjelovi{\'c} and A.~Zisserman, ``{Three things everyone should know to
  improve object retrieval},'' in \emph{2012 IEEE Conference on Computer Vision
  and Pattern Recognition}.\hskip 1em plus 0.5em minus 0.4em\relax IEEE, 2012,
  pp. 2911--2918.

\bibitem{tyszkiewicz2020disk}
M.~Tyszkiewicz, P.~Fua, and E.~Trulls, ``{DISK: Learning local features with
  policy gradient},'' \emph{Advances in Neural Information Processing Systems},
  vol.~33, pp. 14\,254--14\,265, 2020.

\bibitem{Krizhevsky2012}
A.~Krizhevsky, I.~Sutskever, and G.~E. Hinton, ``{ImageNet Classification with
  Deep Convolutional Neural Networks},'' \emph{Advances In Neural Information
  Processing Systems (NeurIPS)}, pp. 1--9, 2012.

\bibitem{Balntas2017HPatches:Descriptors}
V.~Balntas, K.~Lenc, A.~Vedaldi, and K.~Mikolajczyk, ``{HPatches: A benchmark
  and evaluation of handcrafted and learned local descriptors},'' in
  \emph{Proceedings of the IEEE Conference on Computer Vision and Pattern
  Recognition (CVPR)}, 2017.

\bibitem{Schonberger2017ComparativeFeatures}
J.~L. Sch{\"{o}}nberger, H.~Hardmeier, T.~Sattler, and M.~Pollefeys,
  ``{Comparative Evaluation of Hand-Crafted and Learned Local Features},'' in
  \emph{Proceedings of the IEEE Conference on Computer Vision and Pattern
  Recognition (CVPR)}, 2017.

\bibitem{moreels2007evaluation}
P.~Moreels and P.~Perona, ``{Evaluation of features detectors and descriptors
  based on 3d objects},'' \emph{International journal of computer vision},
  vol.~73, no.~3, pp. 263--284, 2007.

\bibitem{bellavia2020there}
F.~Bellavia and C.~Colombo, ``{Is there anything new to say about sift
  matching?}'' \emph{International Journal of Computer Vision}, vol. 128,
  no.~7, pp. 1847--1866, 2020.

\bibitem{laptev2005space}
I.~Laptev, ``{On space-time interest points},'' \emph{International journal of
  computer vision}, vol.~64, no.~2, pp. 107--123, 2005.

\bibitem{scovanner20073}
P.~Scovanner, S.~Ali, and M.~Shah, ``{A 3-dimensional sift descriptor and its
  application to action recognition},'' in \emph{Proceedings of the 15th ACM
  international conference on Multimedia}, 2007, pp. 357--360.

\bibitem{cheung2009n}
W.~Cheung and G.~Hamarneh, ``{$ n $-{SIFT}: $ n $-Dimensional Scale Invariant
  Feature Transform},'' \emph{IEEE Transactions on Image Processing}, vol.~18,
  no.~9, pp. 2012--2021, 2009.

\bibitem{allaire2008full}
S.~Allaire, J.~J. Kim, S.~L. Breen, D.~A. Jaffray, and V.~Pekar, ``{Full
  orientation invariance and improved feature selectivity of 3D SIFT with
  application to medical image analysis},'' in \emph{2008 IEEE computer society
  conference on computer vision and pattern recognition workshops}.\hskip 1em
  plus 0.5em minus 0.4em\relax IEEE, 2008, pp. 1--8.

\bibitem{bersvendsen2016robust}
J.~Bersvendsen \emph{et~al.}, ``{Robust spatio-temporal registration of 4D
  cardiac ultrasound sequences},'' in \emph{Medical Imaging 2016: Ultrasonic
  Imaging and Tomography}, vol. 9790.\hskip 1em plus 0.5em minus 0.4em\relax
  SPIE, 2016, pp. 122--128.

\bibitem{ni2008volumetric}
D.~Ni \emph{et~al.}, ``{Volumetric ultrasound panorama based on 3D SIFT},'' in
  \emph{International conference on medical image computing and
  computer-assisted intervention}.\hskip 1em plus 0.5em minus 0.4em\relax
  Springer, 2008, pp. 52--60.

\bibitem{Toews2016}
M.~Toews and W.~Wells, ``{How are siblings similar? How similar are siblings?
  Large-scale imaging genetics using local image features},'' in
  \emph{International Symposium on Biomedical Imaging (ISBI)}.\hskip 1em plus
  0.5em minus 0.4em\relax IEEE, 4 2016, pp. 847--850.

\bibitem{Kumar2018}
K.~Kumar, L.~Chauvin, M.~Toews, O.~Colliot, and C.~Desrosiers, ``{Multi-Modal
  Analysis of Genetically-Related Subjects Using SIFT Descriptors in Brain
  MRI},'' in \emph{Computational Diffusion MRI}.\hskip 1em plus 0.5em minus
  0.4em\relax Springer, Cham, 2018, pp. 219--228.

\bibitem{Chauvin2019AnalyzingManifold}
L.~Chauvin, K.~Kumar, C.~Desrosiers, J.~De~Guise, W.~Wells, and M.~Toews,
  ``{Analyzing Brain Morphology on the Bag-of-Features Manifold},'' in
  \emph{International Conference on Information Processing in Medical Imaging
  (IPMI)}, vol. 11492 LNCS.\hskip 1em plus 0.5em minus 0.4em\relax Springer
  Verlag, 2019, pp. 45--56.

\bibitem{Chauvin2020NeuroimageRelatives}
L.~Chauvin \emph{et~al.}, ``{Neuroimage signature from salient keypoints is
  highly specific to individuals and shared by close relatives},''
  \emph{NeuroImage}, vol. 204, no.~20, 2020.

\bibitem{pepin2020large}
{\'E}.~Pepin, J.-B. Carluer, L.~Chauvin, M.~Toews, and R.~Harmouche,
  ``{Large-Scale unbiased neuroimage indexing via 3D GPU-SIFT filtering and
  keypoint masking},'' in \emph{Machine Learning in Clinical Neuroimaging and
  Radiogenomics in Neuro-oncology}.\hskip 1em plus 0.5em minus 0.4em\relax
  Springer, 2020, pp. 108--118.

\bibitem{Saiti2020AnMethods}
E.~Saiti and T.~Theoharis, ``{An application independent review of multimodal
  3D registration methods},'' \emph{Computers {\&} Graphics}, vol.~91, pp.
  153--178, 10 2020.

\bibitem{pomerleau2015review}
F.~Pomerleau, F.~Colas, and R.~Siegwart, ``{A review of point cloud
  registration algorithms for mobile robotics},'' \emph{Foundations and Trends
  in Robotics}, vol.~4, no.~1, pp. 1--104, 2015.

\bibitem{besl1992method}
P.~J. Besl and N.~D. McKay, ``{Method for registration of 3-D shapes},'' in
  \emph{Sensor fusion IV: control paradigms and data structures}, vol.
  1611.\hskip 1em plus 0.5em minus 0.4em\relax Spie, 1992, pp. 586--606.

\bibitem{machado2018non}
I.~Machado \emph{et~al.}, ``{Non-rigid registration of 3D ultrasound for
  neurosurgery using automatic feature detection and matching},''
  \emph{International journal of computer assisted radiology and surgery},
  vol.~13, no.~10, pp. 1525--1538, 2018.

\bibitem{frisken2019preliminary}
S.~Frisken \emph{et~al.}, ``{Preliminary results comparing thin-plate splines
  with finite element methods for modeling brain deformation during
  neurosurgery using intraoperative ultrasound},'' in \emph{Medical Imaging
  2019: Image-Guided Procedures, Robotic Interventions, and Modeling}, vol.
  10951.\hskip 1em plus 0.5em minus 0.4em\relax SPIE, 2019, pp. 526--532.

\bibitem{Jiao2019A3D-SIFT}
\BIBentryALTinterwordspacing
Z.~Jiao, R.~Liu, P.~Yi, and D.~Zhou, ``{A Point Cloud Registration Algorithm
  Based on 3D-SIFT},'' \emph{Lecture Notes in Computer Science (including
  subseries Lecture Notes in Artificial Intelligence and Lecture Notes in
  Bioinformatics)}, vol. 11345 LNCS, pp. 24--31, 2019.
\BIBentrySTDinterwordspacing

\bibitem{Golyanik2016GravitationalRegistration}
V.~Golyanik, S.~A. Ali, and D.~Stricker, ``{Gravitational approach for point
  set registration},'' \emph{Proceedings of the IEEE Computer Society
  Conference on Computer Vision and Pattern Recognition}, vol. 2016-December,
  pp. 5802--5810, 12 2016.

\bibitem{Jauer2019EfficientClouds}
P.~Jauer, I.~Kuhlemann, R.~Bruder, A.~Schweikard, and F.~Ernst, ``{Efficient
  registration of high-resolution feature enhanced point clouds},'' \emph{IEEE
  Transactions on Pattern Analysis and Machine Intelligence}, vol.~41, no.~5,
  pp. 1102--1115, 5 2019.

\bibitem{riggi2006fundamental}
F.~Riggi, M.~Toews, and T.~Arbel, ``{Fundamental matrix estimation via
  TIP-transfer of invariant parameters},'' in \emph{18th International
  Conference on Pattern Recognition (ICPR'06)}, vol.~2.\hskip 1em plus 0.5em
  minus 0.4em\relax IEEE, 2006, pp. 21--24.

\bibitem{Ma2017RemoteMatching}
W.~Ma \emph{et~al.}, ``{Remote sensing image registration with modified sift
  and enhanced feature matching},'' \emph{IEEE Geoscience and Remote Sensing
  Letters}, vol.~14, no.~1, pp. 3--7, 1 2017.

\bibitem{arsigny2009fast}
V.~Arsigny, O.~Commowick, N.~Ayache, and X.~Pennec, ``{A fast and log-euclidean
  polyaffine framework for locally linear registration},'' \emph{Journal of
  Mathematical Imaging and Vision}, vol.~33, no.~2, pp. 222--238, 2009.

\bibitem{toews2007statistical}
M.~Toews and T.~Arbel, ``{A statistical parts-based model of anatomical
  variability},'' \emph{IEEE Transactions on Medical Imaging}, vol.~26, no.~4,
  pp. 497--508, 2007.

\bibitem{hu2021end}
J.~Hu \emph{et~al.}, ``{End-to-end multimodal image registration via
  reinforcement learning},'' \emph{Medical Image Analysis}, vol.~68, p. 101878,
  2021.

\bibitem{Viola1997AlignmentInformation}
P.~Viola and W.~M. Wells, ``{Alignment by Maximization of Mutual
  Information},'' \emph{International Journal of Computer Vision}, vol.~24,
  no.~2, pp. 137--154, 1997.

\bibitem{xia2013robust}
R.~Xia, J.~Zhao, and Y.~Liu, ``{A robust feature-based registration method of
  multimodal image using phase congruency and coherent point drift},'' in
  \emph{MIPPR 2013: Pattern Recognition and Computer Vision}, vol. 8919.\hskip
  1em plus 0.5em minus 0.4em\relax International Society for Optics and
  Photonics, 2013, p. 891903.

\bibitem{Wachinger2012EntropyRegistration}
C.~Wachinger and N.~Navab, ``{Entropy and Laplacian images: Structural
  representations for multi-modal registration},'' \emph{Medical Image
  Analysis}, vol.~16, no.~1, pp. 1--17, 1 2012.

\bibitem{kelman2007keypoint}
A.~Kelman, M.~Sofka, and C.~V. Stewart, ``{Keypoint descriptors for matching
  across multiple image modalities and non-linear intensity variations},'' in
  \emph{2007 IEEE conference on computer vision and pattern recognition}.\hskip
  1em plus 0.5em minus 0.4em\relax IEEE, 2007, pp. 1--7.

\bibitem{Bingjian2011ImageImages}
\BIBentryALTinterwordspacing
W.~Bingjian \emph{et~al.}, ``{Image registration method for multimodal
  images},'' \emph{Applied Optics, Vol. 50, Issue 13, pp. 1861-1867}, vol.~50,
  no.~13, pp. 1861--1867, 5 2011.
\BIBentrySTDinterwordspacing

\bibitem{toews2013feature}
M.~Toews, L.~Z{\"{o}}llei, and W.~Wells, ``{Feature-based alignment of
  volumetric multi-modal images},'' in \emph{International Conference on
  Information Processing in Medical Imaging (IPMI)}, vol.~23.\hskip 1em plus
  0.5em minus 0.4em\relax Springer, 2013, pp. 25--36.

\bibitem{noether1971invariant}
E.~Noether, ``{Invariant variation problems},'' \emph{Transport theory and
  statistical physics}, vol.~1, no.~3, pp. 186--207, 1971.

\bibitem{tanaka2021noether}
H.~Tanaka and D.~Kunin, ``Noether’s learning dynamics: Role of symmetry
  breaking in neural networks,'' \emph{Advances in Neural Information
  Processing Systems}, vol.~34, 2021.

\bibitem{florack1992scale}
L.~M. Florack, B.~M. ter Haar~Romeny, J.~J. Koenderink, and M.~A. Viergever,
  ``{Scale and the differential structure of images},'' \emph{Image and vision
  computing}, vol.~10, no.~6, pp. 376--388, 1992.

\bibitem{olver1994differential}
P.~Olver, G.~Sapiro, and A.~Tannenbaum, ``{Differential invariant signatures
  and flows in computer vision: a symmetry group approach},'' in
  \emph{Geometry-driven diffusion in computer vision}.\hskip 1em plus 0.5em
  minus 0.4em\relax Springer, 1994, pp. 255--306.

\bibitem{lecun1989backpropagation}
Y.~LeCun \emph{et~al.}, ``{Backpropagation applied to handwritten zip code
  recognition},'' \emph{Neural computation}, vol.~1, no.~4, pp. 541--551, 1989.

\bibitem{krizhevsky2012imagenet}
A.~Krizhevsky, I.~Sutskever, and G.~E. Hinton, ``{Imagenet classification with
  deep convolutional neural networks},'' \emph{Advances in neural information
  processing systems}, vol.~25, 2012.

\bibitem{bardes2021vicreg}
A.~Bardes, J.~Ponce, and Y.~LeCun, ``{Vicreg: Variance-invariance-covariance
  regularization for self-supervised learning},'' \emph{arXiv preprint
  arXiv:2105.04906}, 2021.

\bibitem{esteves2018learning}
C.~Esteves, C.~Allen-Blanchette, A.~Makadia, and K.~Daniilidis, ``{Learning so
  (3) equivariant representations with spherical cnns},'' in \emph{Proceedings
  of the European Conference on Computer Vision (ECCV)}, 2018, pp. 52--68.

\bibitem{spezialetti2019learning}
R.~Spezialetti, S.~Salti, and L.~D. Stefano, ``{Learning an effective
  equivariant 3D descriptor without supervision},'' in \emph{Proceedings of the
  IEEE/CVF International Conference on Computer Vision}, 2019, pp. 6401--6410.

\bibitem{moyer2021equivariant}
D.~Moyer, E.~Abaci~Turk, P.~E. Grant, W.~M. Wells, and P.~Golland,
  ``{Equivariant Filters for Efficient Tracking in 3D Imaging},'' in
  \emph{International Conference on Medical Image Computing and
  Computer-Assisted Intervention}.\hskip 1em plus 0.5em minus 0.4em\relax
  Springer, 2021, pp. 193--202.

\bibitem{bloem2020probabilistic}
B.~Bloem-Reddy and Y.~W. Teh, ``{Probabilistic Symmetries and Invariant Neural
  Networks.}'' \emph{J. Mach. Learn. Res.}, vol.~21, pp. 90--1, 2020.

\bibitem{dimaio2009spherical}
F.~P. DiMaio, A.~B. Soni, G.~N. Phillips, and J.~W. Shavlik,
  ``{Spherical-harmonic decomposition for molecular recognition in
  electron-density maps},'' \emph{International journal of data mining and
  bioinformatics}, vol.~3, no.~2, pp. 205--227, 2009.

\bibitem{dempster1977maximum}
A.~P. Dempster, N.~M. Laird, and D.~B. Rubin, ``{Maximum likelihood from
  incomplete data via the EM algorithm},'' \emph{Journal of the Royal
  Statistical Society: Series B (Methodological)}, vol.~39, no.~1, pp. 1--22,
  1977.

\bibitem{hough1959machine}
P.~V. Hough, ``{Machine analysis of bubble chamber pictures},'' in \emph{Proc.
  of the International Conference on High Energy Accelerators and
  Instrumentation, Sept. 1959}, 1959, pp. 554--556.

\bibitem{comaniciu2002mean}
D.~Comaniciu and P.~Meer, ``{Mean shift: A robust approach toward feature space
  analysis},'' \emph{IEEE Transactions on pattern analysis and machine
  intelligence}, vol.~24, no.~5, pp. 603--619, 2002.

\bibitem{VanEssen2012}
D.~Van~Essen \emph{et~al.}, ``{The Human Connectome Project: A data acquisition
  perspective},'' \emph{NeuroImage}, vol.~62, no.~4, pp. 2222--2231, 2012.

\bibitem{regan2011genetic}
E.~A. Regan \emph{et~al.}, ``{Genetic epidemiology of COPD (COPDGene) study
  design},'' \emph{COPD: Journal of Chronic Obstructive Pulmonary Disease},
  vol.~7, no.~1, pp. 32--43, 2011.

\bibitem{Menze2015TheBRATS}
\BIBentryALTinterwordspacing
B.~H. Menze \emph{et~al.}, ``{The Multimodal Brain Tumor Image Segmentation
  Benchmark (BRATS)},'' \emph{IEEE transactions on medical imaging}, vol.~34,
  no.~10, pp. 1993--2024, 10 2015.
\BIBentrySTDinterwordspacing

\end{thebibliography}
%

%






\newpage
\appendices

\section{Probabilistic Registration Details}
\label{app:solve}

The probabilistic CPD algorithm is based on a $solve$ function defined for rigid and affine transforms. The $solve\_rigid$ algorithm used in this work is presented in Algorithm~\ref{alg:solve_rigid}, while the $solve\_affine$ can be found in Algorithm~\ref{alg:solve_affine}, from~\cite{myronenko2010point}. The $solve\_rigid()$ function takes as inputs a fixed and a moving point sets $\mathcal{F}$ and $\mathcal{M}$ along with their corresponding probability map $P=\{p_{ij}\}$, and outputs the covariance $\lambda^2$, and a scaled rigid transform with parameters rotation matrix $R$, translation vector $t$, and magnification b. The $solve\_affine()$ function returns an affine transform including invertible affine matrix $B$ and translation vector $t$.

\begin{algorithm}
    \DontPrintSemicolon
    \SetKwFunction{FMain}{solve\_rigid}
    \SetKwProg{Fn}{}{:}{}
    \Fn{\FMain{$\mathcal{F}$, $\mathcal{M}$, $P$}}{
        $\mathit{N}_{\bf{P}}=\bf{1}^{\mathit{T}}\bf{P1},~ \mu_\mathcal{F}=\frac{1}{\mathit{N}_{\bf{P}}}\bm{\mathcal{F}}^{\mathit{T}}\bf{P}^{\mathit{T}}\bf{1},~\mu_\mathcal{M}=\frac{1}{\mathit{N}_{\bf{P}}}\bm{\mathcal{M}}^{\mathit{T}}\bf{P1},$\;
        $\bm{\hat{\mathcal{F}}}=\bm{\mathcal{F}}-\bf{1}\mu_{\mathcal{F}}^{\mathit{T}},~\bm{\hat{\mathcal{M}}}=\bm{\mathcal{M}}-\bf{1}\mu_{\mathcal{M}}^{\mathit{T}},$\;
        $\bf{A}=\bm{\hat{\mathcal{F}}}^{\mathit{T}}\bf{P}^{\mathit{T}}\bm{\hat{\mathcal{M}}}$, compute SVD of $\bf{A}$ such as $\bf{A}=\bf{U}\textit{SS}\bf{V}^{\mathit{T}},$\;
        $\bf{R}=\bf{UCV}^{\mathit{T}}$, where $\bf{C}=\mathrm{d(1,..,1,det(\bf{UV}^{\mathit{T}}))},$\;
        $b=\frac{tr(\bf{A}^{\mathit{T}}\bf{R})}{tr(\bm{\hat{\mathcal{M}}}^{\mathit{T}}d(\bf{P1})\bm{\hat{\mathcal{M}}})},$\;
        $\bf{t}=\mu_{\mathcal{F}}-b\bf{R}\mu_{\mathcal{M}},$\;
        $\lambda^2 = \frac{1}{\mathit{N}_{\bf{P}}D}(tr(\bm{\hat{\mathcal{F}}}^{\mathit{T}}\mathrm{d(\bf{P}^{\mathit{T}}\bf{1})}\bm{\hat{\mathcal{F}}})-\mathnormal{b~tr(\bf{A}^{\mathit{T}}\bf{R}))}.$\;
        \KwRet $\{b,\bf{R},\bf{t}\},\lambda^2$
    }
    \label{alg:solve_rigid}
    \caption{Solve for rigid transform.}
\end{algorithm}

\end{document}


\begin{algorithm}
    \DontPrintSemicolon
    \SetKwFunction{FMain}{solve\_affine}
    \SetKwProg{Fn}{}{:}{}
    \Fn{\FMain{$\mathcal{F}$, $\mathcal{M}$, $P$}}{
    $\mathit{N}_{\bf{P}}=\bf{1}^{\mathit{T}}\bf{P1},~ \mu_\mathcal{F}=\frac{1}{\mathit{N}_{\bf{P}}}\bm{\mathcal{F}}^{\mathit{T}}\bf{P}^{\mathit{T}}\bf{1},~\mu_\mathcal{M}=\frac{1}{\mathit{N}_{\bf{P}}}\bm{\mathcal{M}}^{\mathit{T}}\bf{P1},$\;
    $\bm{\hat{\mathcal{F}}}=\bm{\mathcal{F}}-\bf{1}\mu_{\mathcal{F}}^{\mathit{T}},~\bm{\hat{\mathcal{M}}}=\bm{\mathcal{M}}-\bf{1}\mu_{\mathcal{M}}^{\mathit{T}},$\;
    $\bf{B}=(\bm{\hat{\mathcal{F}}}^{\mathit{T}}\bf{P}^{\mathit{T}}\bm{\hat{\mathcal{M}}})(\bm{\mathcal{M}}^{\mathit{T}}\mathrm{d(\bf{P1})}\bm{\hat{\mathcal{M}}})^{-1},$\;
    $\bf{t}=\mu_{\mathcal{F}}-\bf{B}\mu_{\mathcal{M}},$\;
    $\lambda^2 = \frac{1}{\mathit{N}_{\bf{P}}D}(tr(\bm{\hat{\mathcal{F}}}^{\mathit{T}}\mathrm{d(\bf{P}^{\mathit{T}}\bf{1})}\bm{\hat{\mathcal{F}}})-\mathnormal{tr(\bm{\hat{\mathcal{F}}}^{\mathit{T}}\bf{P}^{\mathit{T}}\bm{\hat{\mathcal{M}}}\bf{B}^{\mathit{T}}))}.$\;
    \KwRet $\{\bf{B},\bf{t}\},\lambda^2$
    }
    \label{alg:solve_affine}
    \caption{Solve for affine transform.}
\end{algorithm}